\documentclass[10pt,twocolumn,letterpaper]{article}

\usepackage[pagenumbers]{cvpr} 
\usepackage[accsupp]{axessibility} 

\usepackage{graphicx}
\usepackage{amsmath}
\usepackage{amssymb}
\usepackage{booktabs}
\usepackage{placeins}

\usepackage[pagebackref,breaklinks,colorlinks]{hyperref}

\usepackage[capitalize]{cleveref}
\crefname{section}{Sec.}{Secs.}
\Crefname{section}{Section}{Sections}
\Crefname{table}{Table}{Tables}
\crefname{table}{Tab.}{Tabs.}

\usepackage{amsmath}
\usepackage{amssymb}
\usepackage{amsthm}

\theoremstyle{definition}

\theoremstyle{remark}

\def\cvprPaperID{7825} 
\def\confName{CVPR}
\def\confYear{2022}

\begin{document}

\title{Deformable ProtoPNet: An Interpretable Image Classifier Using Deformable Prototypes}

\author{Jon Donnelly\\
University of Maine\\
{\tt\small jonathan.donnelly@maine.edu}
\and
Alina Jade Barnett\\
Duke University\\
{\tt\small alina.barnett@duke.edu}
\and
Chaofan Chen\\
University of Maine\\
{\tt\small chaofan.chen@maine.edu}
}
\maketitle

\begin{abstract} 
   We present a deformable prototypical part network (Deformable ProtoPNet), an interpretable image classifier that integrates the power of deep learning and the interpretability of case-based reasoning. This model classifies input images by comparing them with prototypes learned during training, yielding explanations in the form of ``this looks like that.'' However, while previous methods use spatially rigid prototypes, we address this shortcoming by proposing spatially flexible prototypes. Each prototype is made up of several prototypical parts that adaptively change their relative spatial positions depending on the input image. Consequently, a Deformable ProtoPNet can explicitly capture pose variations and context, improving both model accuracy and the richness of explanations provided. Compared to other case-based interpretable models using prototypes, our approach achieves state-of-the-art accuracy and gives an explanation with greater context. The code is available at \href{https://github.com/jdonnelly36/Deformable-ProtoPNet}{https://github.com/jdonnelly36/Deformable-ProtoPNet}.
\end{abstract}

\section{Introduction}
\label{sec:intro}


\begin{figure}[t]
  \centering
    \includegraphics[width=0.4\textwidth]{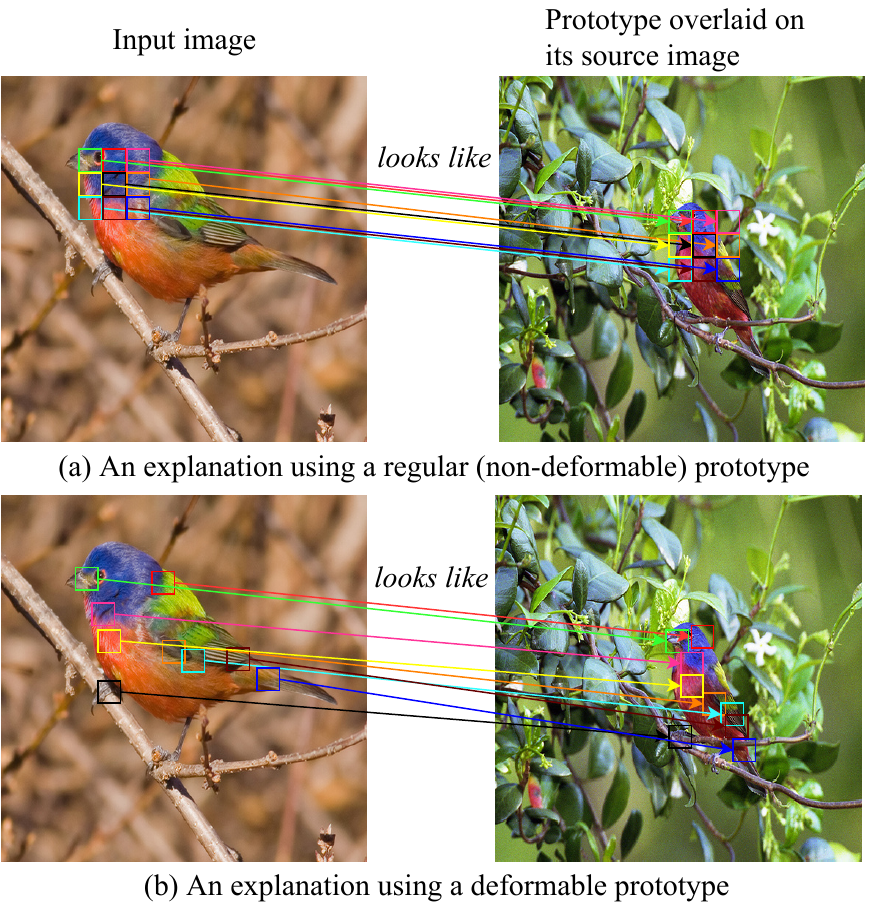}
  \caption{How an input image of a painted bunting is compared with (a) a regular (non-deformable) prototype and (b) a deformable prototype of the painted bunting class (overlaid on its source image).} 
  \label{fig:intro}
\end{figure}

Machine learning has been adopted in many domains, including high-stakes applications such as healthcare \cite{nayyar2021machine,barnett_iaia-bl_2021}, finance \cite{warin2021machine}, and criminal justice \cite{berk2019machine}. In these critical domains, interpretability is essential in determining whether we can trust predictions made by machine learning models. 
In computer vision, there is a growing stream of research that aims to produce accurate yet interpretable image classifiers by integrating the power of deep learning and the interpretability of case-based reasoning \cite{ProtoPNet, ProtoTree, TesNet, barnett_iaia-bl_2021}. These models learn a set of \textit{prototypes} from training images, and make predictions by comparing parts of the input image with prototypes learned during training. This enables explanations of the form ``this is an image of a painted bunting, because \textit{this} part of the image looks like \textit{that} prototypical part of a painted bunting,'' as in Figure \ref{fig:intro}(a). 
However, existing prototype-based models for computer vision use spatially rigid prototypes, which cannot explicitly account for geometric transformations or pose variations of objects.

Inspired by recent work on modeling geometric transformations in convolutional neural networks \cite{jaderberg2015spatial,jeon2017active_conv,dai2017deformable_conv,zhu2019deformable_conv_v2}, we propose a \textit{deformable prototypical part network} (\textit{Deformable ProtoPNet}), a case-based interpretable neural network that provides spatially flexible \textit{deformable prototypes}. In a Deformable ProtoPNet, each prototype is made up of several prototypical parts that \textit{adaptively change their relative spatial positions} depending on the input image. This enables each prototype to detect object features with a higher tolerance to spatial transformations, as the parts within a prototype are allowed to move. Figure \ref{fig:intro}(b) illustrates the idea of a deformable prototype; when an input image is compared with a deformable prototype, the prototypical parts within the deformable prototype adaptively change their relative spatial positions to detect similar parts of the input image. Consequently, a Deformable ProtoPNet can explicitly capture pose variations, and improve both model accuracy and the richness of explanations provided.

The main contributions of our paper are as follows: (1) We developed the first prototypical case-based interpretable neural network that provides spatially flexible deformable prototypes. (2) We improved the accuracy of case-based interpretable neural networks by introducing angular margins to the training algorithm. (3) We showed that Deformable ProtoPNet can achieve state-of-the-art accuracy on the CUB-200-2011 bird recognition dataset \cite{cub_200} and the Stanford Dogs \cite{dogs} dataset.

\section{Related Work}\label{sec:RW}

There are two general approaches to interpreting deep neural networks: (1) explaining trained neural networks \textit{posthoc}; and (2) building inherently interpretable neural networks that can explain themselves. 
\textit{Posthoc} explanation techniques (e.g., model approximations using interpretable surrogates \cite{ribeiro2016should,lundberg2017unified}, activation maximizations \cite{erhan2009visualizing,yosinski2015understanding,nguyen2016synthesizing}, saliency visualizations \cite{simonyan2013deep,zeiler2014visualizing,springenberg2014striving,smilkov2017smoothgrad,sundararajan2017axiomatic,bach2015pixel,shrikumar2017learning}) do not make a neural network \textit{inherently interpretable}, as \textit{posthoc} explanations are not used by the original network during prediction, and may not be faithful to what the original network computes \cite{rudin_stop_2019}.

A Deformable ProtoPNet uses case-based reasoning with prototypes to build an \textit{inherently interpretable network}. 
This idea was explored in \cite{LiLiuChenRudin} and carried further in \cite{ProtoPNet}, where a prototypical part network (ProtoPNet) was introduced. 
ProtoPNet uses the similarity scores between an input image and the learned prototypes to generate explanations for its predictions in the form of ``this looks like that,'' as in Figure \ref{fig:intro}(a). The ProtoPNet model has been extended multiple times \cite{nauta2021looks,ProtoTree,TesNet}. We build our Deformable ProtoPNet upon the ProtoPNet and TesNet models \cite{TesNet}. TesNet uses a cosine similarity metric to compute similarities between image patches and prototypes in a latent space, and introduces loss terms to encourage the prototype vectors within a class to be orthogonal to each other and to separate the latent spaces of different classes.

All previous prototype-based image classifiers use spatially rigid prototypes. In contrast, Deformable ProtoPNet is the first network to use spatially flexible deformable prototypes, where each prototype consists of several prototypical parts that adaptively change their relative spatial positions depending on the input image (Figure \ref{fig:intro}(b)). In this way, our Deformable ProtoPNet can capture pose variations and offer a richer explanation for its predictions than previous image classifiers that use case-based reasoning.

\textbf{Deformations and Geometric Transformations}.
Our work relates closely to previous work modeling object deformations and geometric transformations. One of the early attempts at modeling deformations in computer vision models was provided by Deformable Part Models (DPMs) \cite{felzenszwalb2009object}, which model deformations as deviations of object parts from their (heuristically chosen) ``anchor'' positions. The original DPMs use histogram-of-oriented-gradients (HOG) features \cite{dalal2005histograms} to represent objects and their parts, and are trained using latent support vector machines (SVMs). The idea of modeling spatial deformations, encapsulated in DPMs, has been extended into convolutional neural networks (CNNs). The inference algorithm of a DPM was shown to be equivalent to a CNN with distance transform pooling \cite{girshick2015deformable}, and distance transform pooling was extended into a deformation layer in \cite{ouyang2013joint} for pedestrian detection and deformation pooling (def-pooling) layers in a DeepID-Net \cite{ouyang2015deepid} for generic object detection. More recent developments include spatial transformer networks \cite{jaderberg2015spatial}, and networks with active convolutions \cite{jeon2017active_conv} and deformable convolutions \cite{dai2017deformable_conv,zhu2019deformable_conv_v2}. Spatial transformer networks \cite{jaderberg2015spatial} predict global parametric transformations (e.g., affine transformations) to be applied to an input image or a convolutional feature map, with the goal of normalizing the ``pose'' of the target object in the image. Active convolutions \cite{jeon2017active_conv} learn and apply the same deformation to a convolutional filter, when scanning the filter across all spatial locations of the input feature map. Deformable convolutions \cite{dai2017deformable_conv,zhu2019deformable_conv_v2}, on the other hand, learn to predict deformations that will be applied to convolutional filters at each spatial location of the input feature map. This means that the deformations are different across spatial locations and input images.

Our Deformable ProtoPNet builds upon the Deformable Convolutional Network \cite{dai2017deformable_conv,zhu2019deformable_conv_v2}, by using a similar mechanism to generate offsets for deforming prototypes. However, our Deformable ProtoPNet is different from the Deformable Convolutional Network (and the previous work) in two major ways: (1) our Deformable ProtoPNet offers deformable prototypes whose individual parts can be visualized and understood by human beings; (2) by constraining the image features and the representations of prototypes and prototypical parts to be fixed-length vectors, our Deformable ProtoPNet learns an embedding space that has a geometric interpretation, where image features are clustered around similar prototypes on a hypersphere.

\textbf{Deep Metric Learning Using Margins}.
Our work also relates to previous work that performs deep metric learning using cosine \cite{wang2018cosface} or angular margins \cite{liu2017sphereface,liu2016large,wang2018additive,deng2019arcface}. These techniques use unit vectors to represent classes in the fully-connected last layer of a neural network, allowing us to re-interpret the logit of a class for a given input as the cosine of the angle between the class vector and the latent representation of the input. A margin can then be introduced to increase the angle between the latent representation of a training example and the vector of its target class during training, decreasing the target class logit during training, forcing the network to ``try harder'' to further reduce the angle in order to lower the cross entropy loss. At the end of training with margins, the latent representations of the training examples from the same class will be clustered in angular space around the vector of that class, and they will be separated by some angular margin from the latent representations of the training examples from a different class. In training our Deformable ProtoPNet, we apply angular margins to inflate the prototype activations of the incorrect-class prototypes for each training example during training. 



\section{Deformable Prototypes}
\subsection{Overview of a Deformable Prototype}
We will first discuss the general formulation of a non-deformable prototype, as defined in previous work (e.g., \cite{ProtoPNet}). Let $\mathbf{p}^{(c,l)}$ denote the $l$-th prototype of class $c$, represented as a tensor of the shape $\rho_1 \times \rho_2 \times d$ with $\rho = \rho_1\rho_2$ spatial positions, and let $\mathbf{p}^{(c,l)}_{m,n}$ denote the $d$-dimensional vector at the spatial location $(m, n)$ of the prototype tensor $\mathbf{p}^{(c,l)}$, with $m \in \{-\lfloor\rho_1/2\rfloor, ... , \lfloor\rho_1/2\rfloor\}$ and $n \in \{-\lfloor\rho_2/2\rfloor, ... , \lfloor\rho_2/2\rfloor\}$. (A $3 \times 3$ prototype has $\rho_1 = \rho_2 = 3$ and $m, n \in \{-1, 0, 1\}$.) Let $\mathbf{z}$ denote a tensor of image features with shape $\eta_1 \times \eta_2 \times d$, produced by passing an input image through some feature extractor (e.g., a CNN), and let $\mathbf{z}_{a,b}$ denote the $d$-dimensional vector at the spatial location $(a, b)$ of the image-feature tensor $\mathbf{z}$. In the previous work \cite{ProtoPNet}, the prototype's height and the width satisfy $\rho_1 \leq \eta_1$ and $\rho_2 \leq \eta_2$, and its depth is the same as that of $\mathbf{z}$. 
We can interpret each prototype as representing a patch in the input image, and we can compare a prototype with each $\rho_1 \times \rho_2$ patch of an image-feature tensor using an $L^2$-based similarity function. Mathematically, for each spatial position $(a, b)$ in an image-feature tensor $\mathbf{z}$, a regular non-deformable prototype computes its similarity with a $\rho_1 \times \rho_2$ patch of $\mathbf{z}$ centered at $(a, b)$ as:
\begin{equation}\label{eq:general_similarity}
g(\mathbf{z})^{(c,l)}_{a,b}=\text{sim}\left(\sum_{m}\sum_{n}\|\mathbf{p}^{(c,l)}_{m,n} - \mathbf{z}_{a+m,b+n}\|_2^2\right),
\end{equation}
where $\text{sim}$ is a function that inverts an $L^2$-distance (in the latent space of image features) into a similarity measure. In a ProtoPNet \cite{ProtoPNet} and a ProtoTree \cite{ProtoTree}, an $L^2$-based similarity was used to compare a prototype and an image patch in the latent space, presumably because it is intuitive to think about similarity as ``closeness'' in a Euclidean space.

\begin{figure}[t]
  \centering
    \includegraphics[width=0.4\textwidth]{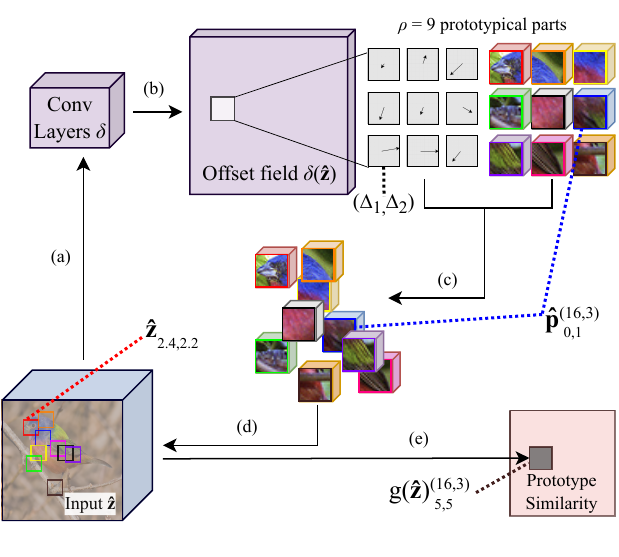}
  \caption{How a deformable prototype is applied to the latent representation of an input image of a painted bunting. (a) The latent input $\mathbf{\hat{z}}$ is fed into the offset prediction function $\delta$ to produce (b) a field of offsets. These offsets are used to (c) alter the spatial position of each prototypical part, which are (d) compared to the input to (e) compute prototype similarity according to equation (\ref{eq:deform_similarity}). }
  \label{fig:deformable_prototypes}
\end{figure}

In a ProtoPNet \cite{ProtoPNet}, a prototype (prototypical part) is a spatially contiguous patch, regardless of the number of its spatial positions $\rho$. For example, Figure \ref{fig:intro}(a)(right) illustrates a $3 \times 3$ non-deformable prototype that can be used in a ProtoPNet. 
In a Deformable ProtoPNet, we define a prototypical part within a (deformable) prototype to be a $1 \times 1$ patch (of shape $1 \times 1 \times d$) within a prototype tensor (of shape $\rho_1 \times \rho_2 \times d$) (see Figure \ref{fig:deformable_prototypes}). In particular, we use $\hat{\mathbf{p}}^{(c,l)}$ to denote the $l$-th deformable prototype of class $c$, again represented as a tensor of the shape $\rho_1 \times \rho_2 \times d$ with $\rho = \rho_1\rho_2$ spatial positions, and we use $\hat{\mathbf{p}}^{(c,l)}_{m,n}$ to denote the $(m, n)$-th prototypical part within the deformable prototype $\hat{\mathbf{p}}^{(c,l)}$.
Figure \ref{fig:intro}(b)(right) illustrates a deformable prototype of $9$ spatial positions (represented as a $3 \times 3 \times d$ tensor), where each spatial position is viewed as an individual prototypical part that can move around, and represents a semantic concept that is \textit{spatially decoupled} from other prototypical parts. For notational consistency, we use $\hat{\mathbf{z}}$ to denote a tensor of image features that will be compared with a deformable prototype $\hat{\mathbf{p}}^{(c,l)}$, and we use $\hat{\mathbf{z}}_{a,b}$ to denote the $d$-dimensional vector at the spatial location $(a, b)$ of the image-feature tensor $\hat{\mathbf{z}}$.

In a Deformable ProtoPNet, we require all prototypical parts $\hat{\mathbf{p}}^{(c,l)}_{m,n}$ (a $d$-dimensional vector) of all deformable prototypes $\hat{\mathbf{p}}^{(c,l)}$ to have the same $L^2$ length:
\begin{equation}\label{eq:fix_length_p}
\|\hat{\mathbf{p}}^{(c,l)}_{m,n}\|_2 = r = 1/\sqrt{\rho},
\end{equation}
so that when we represent a deformable prototype $\hat{\mathbf{p}}^{(c,l)}$ as a stacked vector of its constituent prototypical parts $\hat{\mathbf{p}}^{(c,l)}_{m,n}$, 
all deformable prototypes have the same $L^2$ length, which is equal to $\|\hat{\mathbf{p}}^{(c,l)}\|_2 = \sqrt{\rho r^2} = 1$ (i.e., all deformable prototypes are unit vectors). We also require every spatial location $(a, b)$ of every image-feature tensor $\hat{\mathbf{z}}$ to have the same $L^2$ length:
\begin{equation}\label{eq:fix_length_z}
\|\hat{\mathbf{z}}_{a,b}\|_2 = r = 1/\sqrt{\rho}.
\end{equation}
With equations (\ref{eq:fix_length_p}) and (\ref{eq:fix_length_z}), we can rewrite the squared $L^2$ distance between $\hat{\mathbf{p}}^{(c,l)}_{m,n}$ and $\hat{\mathbf{z}}_{a+m,b+n}$ in equation (\ref{eq:general_similarity}) as:
$\|\hat{\mathbf{p}}^{(c,l)}_{m,n} - \hat{\mathbf{z}}_{a+m,b+n}\|_2^2 = \sum_{m}\sum_{n} (2r^2 - 2\mathbf{\hat{p}}^{(c,l)}_{m,n} \cdot \mathbf{\hat{z}}_{a+m,b+n})$. With similarity function
$\text{sim}(\kappa)=-(\kappa/2 - 1)$,
the similarity (defined in equation (\ref{eq:general_similarity})) between a deformable prototype $\hat{\mathbf{p}}^{(c,l)}$ of shape $\rho_1 \times \rho_2 \times d$ and a $\rho_1 \times \rho_2$ patch, centered at $(a, b)$, of the image-feature tensor $\hat{\mathbf{z}}$ becomes: 
\begin{equation}\label{eq:pre-deform_similarity}
g(\hat{\mathbf{z}})^{(c,l)}_{a,b} = \sum_{m}\sum_{n} \mathbf{\hat{p}}^{(c,l)}_{m,n} \cdot \mathbf{\hat{z}}_{a+m,b+n}
\end{equation}
\textit{before} we allow the prototype to deform. Note that equation (\ref{eq:pre-deform_similarity}) is equivalent to a convolution between $\hat{\mathbf{p}}^{(c,l)}$ and $\hat{\mathbf{z}}$, but with the added constraints given by equations (\ref{eq:fix_length_p}) and (\ref{eq:fix_length_z}).

To allow a deformable prototype $\hat{\mathbf{p}}^{(c,l)}$ to deform, we introduce offsets to enable each prototypical part $\hat{\mathbf{p}}^{(c,l)}_{m,n}$ to move around when the prototype is applied at a spatial location $(a, b)$ on the image-feature tensor $\hat{\mathbf{z}}$. Mathematically, equation (\ref{eq:pre-deform_similarity}) becomes:
\begin{equation}\label{eq:deform_similarity}
g(\hat{\mathbf{z}})^{(c,l)}_{a,b} = \sum_{m}\sum_{n} \mathbf{\hat{p}}^{(c,l)}_{m,n} \cdot \mathbf{\hat{z}}_{a+m+\Delta_1,b+n+\Delta_2},
\end{equation}
where $\Delta_1 = \Delta_1(\mathbf{\hat{z}}, a, b, m, n)$ and $\Delta_2 = \Delta_2(\mathbf{\hat{z}}, a, b, m, n)$ are functions depending on $\mathbf{\hat{z}}$, $a$, $b$, $m$, and $n$ 
(further explained in Section \ref{sec:offset_generation}). These offsets allow us to evaluate the similarity between a prototypical part $\hat{\mathbf{p}}^{(c,l)}_{m,n}$ and the image feature $\mathbf{\hat{z}}_{a+m+\Delta_1,b+n+\Delta_2}$ at a deformed position $(a+m+\Delta_1,b+n+\Delta_2)$ rather than the regular grid position $(a+m, b+n)$. 
Since $\Delta_1$ and $\Delta_2$ are typically fractional, we use feature interpolation to define image features at fractional positions (discussed in Section \ref{sec:offset_generation}). 
We further require interpolated image features to have the same $L^2$ length of $r$ as those image features at regular grid positions, namely:
\begin{equation}\label{eq:fix_length_interp}
\|\mathbf{\hat{z}}_{a+m+\Delta_1,b+n+\Delta_2}\|_2 = r = 1/\sqrt{\rho}.
\end{equation}
Note that equation (\ref{eq:deform_similarity}) is equivalent to a deformable convolution \cite{dai2017deformable_conv, zhu2019deformable_conv_v2} between $\hat{\mathbf{p}}^{(c,l)}$ and $\hat{\mathbf{z}}$, but with the added constraints given by equations (\ref{eq:fix_length_p}) and (\ref{eq:fix_length_interp}).

It is worth noting that similarity defined in equation (\ref{eq:deform_similarity}) has a simple geometric interpretation. Let $\theta(\mathbf{v},\mathbf{w})$ denote the angle between two vectors, and let 
\begin{equation}
\label{eq:def_partwise_similarity}
    g(\hat{\mathbf{z}})^{(c,l)}_{a,b,m,n} = \mathbf{\hat{p}}^{(c,l)}_{m,n} \cdot \mathbf{\hat{z}}_{a+m+\Delta_1,b+n+\Delta_2}
\end{equation} 
denote the contribution of the prototypical part $\hat{\mathbf{p}}^{(c,l)}_{m,n}$ to the similarity score of the prototype. Note that equation (\ref{eq:def_partwise_similarity}) is \textit{exactly} equal to $\cos{(\theta(\mathbf{\hat{p}}^{(c,l)}_{m,n}, \mathbf{\hat{z}}_{a+m+\Delta_1,b+n+\Delta_2}))}$, which is the cosine similarity between $\mathbf{\hat{p}}^{(c,l)}_{m,n}$ and  $\mathbf{\hat{z}}_{a+m+\Delta_1,b+n+\Delta_2}$. Since both $\hat{\mathbf{p}}^{(c,l)}_{m,n}$ and $\hat{\mathbf{z}}_{a+m+\Delta_1,b+n+\Delta_2}$ have the same $L^2$ length $r$ (equations (\ref{eq:fix_length_p}) and (\ref{eq:fix_length_interp})), all prototypical parts and all (interpolated) image features live on a $d$-dimensional hypersphere of radius $r$. This means that an interpolated image feature vector $\mathbf{\hat{z}}_{a+m+\Delta_1,b+n+\Delta_2}$ is considered similar (has a large cosine similarity) to a prototypical part $\mathbf{\hat{p}}^{(c,l)}_{m,n}$ \textit{only} when the angle between them is small on the hypersphere. 

A similar geometric interpretation also holds between an entire deformable prototype and image features at deformed positions. Let $\hat{\mathbf{z}}^\Delta_{a,b}$
denote the interpolated image features $\hat{\mathbf{z}}_{a-\lfloor\rho_1/2\rfloor+\Delta_1,b-\lfloor\rho_2/2\rfloor+\Delta_2}$, ..., $\hat{\mathbf{z}}_{a+\lfloor\rho_1/2\rfloor+\Delta_1,b+\lfloor\rho_2/2\rfloor+\Delta_2}$ at $\rho$ deformed positions, stacked into a column vector. Note that $\hat{\mathbf{z}}^\Delta_{a,b}$ has $L^2$ length $\|\hat{\mathbf{z}}^\Delta_{a,b}\|_2 = 1$. We can then rewrite equation (\ref{eq:deform_similarity}) as:
\[
g(\hat{\mathbf{z}})^{(c,l)}_{a,b} =
\hat{\mathbf{p}}^{(c,l)} \cdot \hat{\mathbf{z}}^\Delta _{a,b} = \cos(\theta(\hat{\mathbf{p}}^{(c,l)}, \hat{\mathbf{z}}^\Delta_{a,b})),
\]
which is exactly the cosine similarity between $\hat{\mathbf{p}}^{(c,l)}$ and $\hat{\mathbf{z}}^\Delta_{a,b}$. Since both $\hat{\mathbf{p}}^{(c,l)}$ and $\hat{\mathbf{z}}^\Delta_{a,b}$ are unit vectors, all deformable prototypes and all collections of interpolated image features at $\rho$ deformed positions live on a $\rho d$-dimensional hypersphere of radius $1$. This means that a collection of interpolated image features $\hat{\mathbf{z}}^\Delta_{a,b}$ is considered similar to an \textit{entire} deformable prototype $\hat{\mathbf{p}}^{(c,l)}$ \textit{only} when the angle between them is small on the hypersphere.

With the similarity between a deformable prototype and a collection of image features at deformed locations defined in equation (\ref{eq:deform_similarity}), we now define the similarity score between a deformable prototype $\hat{\mathbf{p}}^{(c,l)}$ and an \textit{entire} image-feature tensor $\hat{\mathbf{z}}$ to be its maximum similarity to any set of positions:
\begin{equation}
\label{eq:overall_similarity}
    g(\hat{\mathbf{z}})^{(c,l)} = \max_{a,b} g(\hat{\mathbf{z}})^{(c,l)}_{a,b}
\end{equation}

In our experiments, we trained Deformable ProtoPNets using both $3 \times 3$ and $2 \times 2$ deformable prototypes. A $2 \times 2$ deformable prototype $\hat{\mathbf{p}}^{(c,l)}$ can be implemented as a tensor of the shape $2 \times 2 \times d$ ($\rho_1 = \rho_2 = 2$) \underline{with dilation $2$} and with $\rho = \rho_1\rho_2 = 4$ prototypical parts at $(m, n) \in \{(-1, -1), (-1, 1), (1, -1), (1, 1)\}$.

\subsection{Offset Generation and Feature Interpolation}
\label{sec:offset_generation}
As in Figure \ref{fig:deformable_prototypes}, the offsets used for deformable prototypes are computed using an offset prediction function $\delta$ that maps fixed length input features $\mathbf{\hat{z}}$ to an offset field with the same spatial size as $\mathbf{\hat{z}}$. At each spatial center location, this field contains $2\rho$ components, corresponding to a $(\Delta_1, \Delta_2)$ pair of offsets for each of the $\rho$ prototypical parts. 


The offsets $(\Delta_1, \Delta_2)$ produced by $\delta$ may be integer or fractional. Prior work \cite{dai2017deformable_conv, zhu2019deformable_conv_v2, jeon2017active_conv, jaderberg2015spatial} uses bilinear interpolation to compute the value of these fractional locations. In contrast, we do not use bilinear interpolation because it is not feasible for a Deformable ProtoPNet, as the similarity function specified in equation (\ref{eq:deform_similarity}) relies on the assumption that the image feature vector $\hat{\mathbf{z}}_{a+m+\Delta_1,b+n+\Delta_2}$ is of $L^2$ length $r$; without this assumption, similarities will no longer be dependent only on the angle between a prototype and image features. Bilinear interpolation breaks this assumption, because when interpolating between two vectors that have the same $L^2$ norm, bilinear interpolation does not preserve the $L^2$ norm for the interpolated vector. This can be informally explained geometrically: bilinear interpolation chooses a point on the hyperplane that intersects the four interpolated points, meaning that it will never fall on the hypersphere for a fractional location.  
We use an $L^2$ norm-preserving interpolation function, introduced in Theorem 3.1, to solve this problem. A proof of Theorem 3.1 can be found in the supplement. 

\textbf{Theorem 3.1.}\label{thm.bilinear_squared} Let $\mathbf{\hat{z}}_1, \mathbf{\hat{z}}_2, \mathbf{\hat{z}}_3, \mathbf{\hat{z}}_4 \in \mathbb{R}^n$ be vectors such that $\|\mathbf{\hat{z}}_i\| = r$ for all $i \in {1,2,3,4}$ for some constant $r$, and let $\hat{\mathbf{z}}^2$ denote the element-wise square of a vector. For some constants $\alpha \in [0, 1]$ and $\beta \in [0,1]$, the bilinear interpolation operation $\mathbf{z}_{\text{interp}}=(1-\alpha)(1-\beta)\mathbf{\hat{z}}_1 + (1 - \alpha)\beta\mathbf{\hat{z}}_2 + \alpha(1-\beta)\mathbf{\hat{z}}_3 + \alpha\beta\mathbf{\hat{z}}_4$ does not guarantee that $\|\mathbf{z}_{\text{interp}}\|_2 = r$. However, the $L^2$ norm-preserving interpolation operation $\mathbf{z}_{\text{interp}}=\sqrt{(1-\alpha)(1-\beta)\mathbf{\hat{z}}^2_1 + (1 - \alpha)\beta\mathbf{\hat{z}}^2_2 + \alpha(1-\beta)\mathbf{\hat{z}}^2_3 + \alpha\beta\mathbf{\hat{z}}^2_4}$ guarantees that $\|\mathbf{z}_{\text{interp}}\|_2 = r$.

\label{sec:deformable_prototypes:implementation}
Finally, the theoretical framework of a deformable prototype requires that every spatial location $\hat{\mathbf{z}}_{a,b}$ and $\mathbf{\hat{p}}^{(c,l)}_{m,n}$ of $\hat{\mathbf{z}}$ and $\hat{\mathbf{p}}^{(c,l)}$ have $L^2$ length $r$. In our implementation, we guarantee this by always normalizing and scaling both the image features extracted by a CNN at every spatial location $(a,b)$ of a convolutional output $\mathbf{z}$, as well as every prototypical part of a deformable prototype, to length $r$, before they are used in computation. Specifically, we compute $\mathbf{\hat{z}}_{a,b} = r \mathbf{z}_{a,b}/\|\mathbf{z}_{a,b}\|_2$ for every spatial location $(a,b)$ of the convolutional output $\mathbf{z}$ and $\mathbf{\hat{p}}^{(c,l)}_{m,n} = r \mathbf{p}_{m,n}^{(c,l)}/\|\mathbf{p}_{m,n}^{(c,l)}\|_2$ for every $(m,n)$-th part of a deformable prototype. However, this normalization is undefined when $\|\mathbf{p}^{(c,l)}_{m,n}\|_2=0$ or $\|\mathbf{z}_{a,b}\|_2=0$. This is a problem because zero padding and the ReLU activation function can both create a feature vector $\mathbf{z}$ with $L^2$ norm 0. We address this problem by appending a uniform channel of a small value $\epsilon=10^{-5}$ to $\mathbf{p}^{(c,l)}$ and $\mathbf{z}$ prior to normalization. In particular, an all-$0$ feature vector $\mathbf{z}_{a,b}$ produced by a CNN will become $\begin{bmatrix}
0 ~ \hdots ~ 0 ~ \epsilon
\end{bmatrix}$, which has an $L^2$ norm of $\epsilon$. 



\section{Deformable ProtoPNet}

\begin{figure*}
    \centering
    \includegraphics[width=0.9\textwidth]{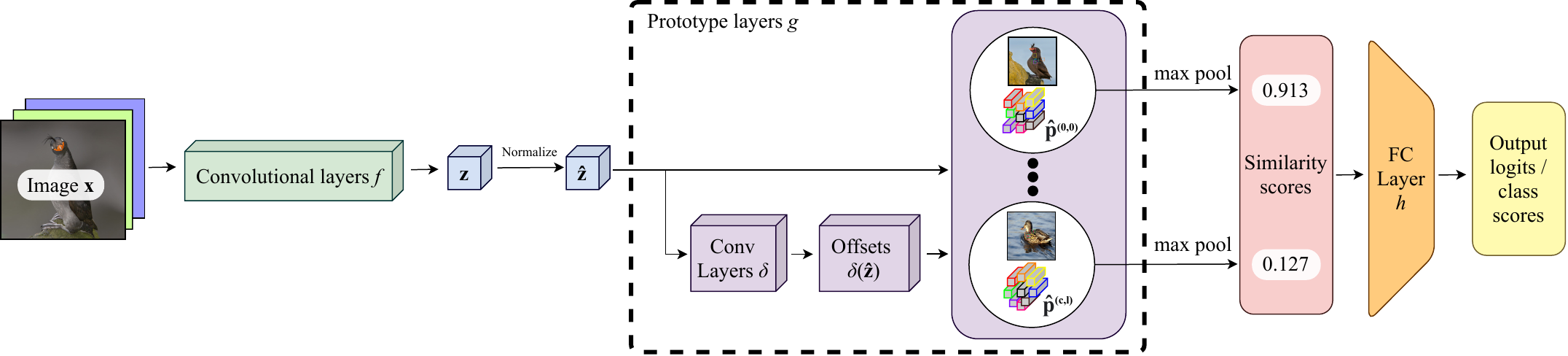}
\captionsetup{width=0.9\textwidth}
  \caption{The architecture for Deformable ProtoPNet. 
  }
  \label{fig:architecture}
\end{figure*}

Figure \ref{fig:architecture} gives an overview of the architecture of a Deformable ProtoPNet. A Deformable ProtoPNet consists of a CNN backbone $f$ that maps an image $\mathbf{x}$ to latent image features $\mathbf{z}$, which are normalized to length $r$ at each spatial location into $\hat{\mathbf{z}}$, followed by a deformable prototype layer $g$ that contains deformable prototypes as defined in Section \ref{sec:deformable_prototypes:implementation}, and a fully connected last layer $h$, which combines the similarity scores produced by deformable prototypes into a class score for each class. 

\subsection{Training}
Similar to \cite{ProtoPNet}, the training of a Deformable ProtoPNet proceeds in three stages. 

\textbf{Stage 1: Stochastic gradient descent (SGD) of layers before last layer.} We perform stochastic gradient descent over the features of $f$ and $g$ while keeping $h$ fixed. By doing so we aim to learn a useful feature space where the image features $\mathbf{\hat{z}}^{\Delta}_{a,b}$ of inputs of class $c$ are clustered around prototypes $\mathbf{\hat{p}}^{(c,l)}$ of the same class, but separated from those of other classes on a hypersphere. To achieve this, we use the cluster and separation losses as in \cite{ProtoPNet} and adapted for the angular space in \cite{TesNet}. The cluster and separation losses are defined as:
\begin{equation}
\label{eqn:cluster}
    \ell_{\text{clst}} = -\frac{1}{N} \sum_{i=1}^N \max_{\mathbf{\hat{p}}^{(c,l)}:c= y^{(i)}} 
    g(\mathbf{\hat{z}}^{(i)})^{(c,l)}
\end{equation}
and
\begin{equation}
\label{eqn:separation}
    \ell_{\text{sep}} = \frac{1}{N} \sum_{i=1}^N \max_{\mathbf{\hat{p}}^{(c,l)}:c\neq y^{(i)}} 
    g(\mathbf{\hat{z}}^{(i)})^{(c,l)}
\end{equation}
respectively, where $N$ is the total number of inputs, $\mathbf{\hat{z}}^{(i)}$ is the image feature tensor normalized and scaled at each spatial location for input $i$, $y^{(i)}$ is the label of $\mathbf{x}^{(i)}$, and all other values are as defined previously.

We were inspired by recent work in margin-based softmax losses \cite{wang2018cosface,liu2017sphereface,liu2016large,wang2018additive,deng2019arcface} to further encourage this clustering structure by modifying traditional cross entropy loss. Specifically, we use a new form of cross entropy: \textit{subtractive margin cross entropy}. This is defined as:
\begin{equation}
\label{eqn:subtractive_ce}
    \text{CE}^{(-)} = 
    \sum_{i=1}^N
    -\log\frac{\exp{(\sum_{c, l}w_h^{((c,l),y^{(i)})} g^{(-)}(i)^{(c,l)})}}
    { \sum_{c'}\exp{(\sum_{c,l}w_h^{((c,l),c')} g^{(-)}(i)^{(c,l)})} },
\end{equation}
where $w^{((c,l), c')}_h$ denotes the last layer connection between the similarity of prototype $\mathbf{\hat{p}}^{(c,l)}$ and class $c'$,
\begin{equation}
\label{eqn:subtractive_margin}
    g^{(-)}(i)^{(c,l)} = 
    \begin{cases}
    g(\mathbf{\hat{z}}^{(i)})^{(c,l)} \hspace{30mm}\text{if } c = y^{(i)}&\\
    \underset{a,b}{\max}\cos(\lfloor\theta(\mathbf{\hat{p}}^{(c,l)},\mathbf{\hat{z}}^{\Delta,(i)}_{a,b}) - \phi\rfloor_+) \hspace{5mm}\text{else} &
    \end{cases}
\end{equation}
for a fixed margin $\phi=0.1$, and $\lfloor \hspace{1mm}  \rfloor_+$ denotes the ReLU function. Subtractive margin cross entropy encourages a well separated feature space by artificially decreasing the angle between a deformable prototype $\hat{\mathbf{p}}^{(c,l)}$ of class $c$ and the collection of deformed image features $\hat{\mathbf{z}}^{\Delta,(i)}_{a,b}$ from the $i$-th training image with $y^{(i)} \neq c$, thereby inflating the cosine similarity between the two and increasing the class score of the incorrect class $c$. In order to reduce the value of this loss, the network has to try harder to counter the introduced margin $\phi$ by further increasing the angle between a deformable prototype $\hat{\mathbf{p}}^{(c,l)}$ and an image feature $\hat{\mathbf{z}}^{\Delta,(i)}_{a,b}$ of an incorrect class, resulting in a stronger separation between classes on the latent hypersphere.

\begin{figure}[t]
    \centering
    \includegraphics[width=0.45\textwidth]{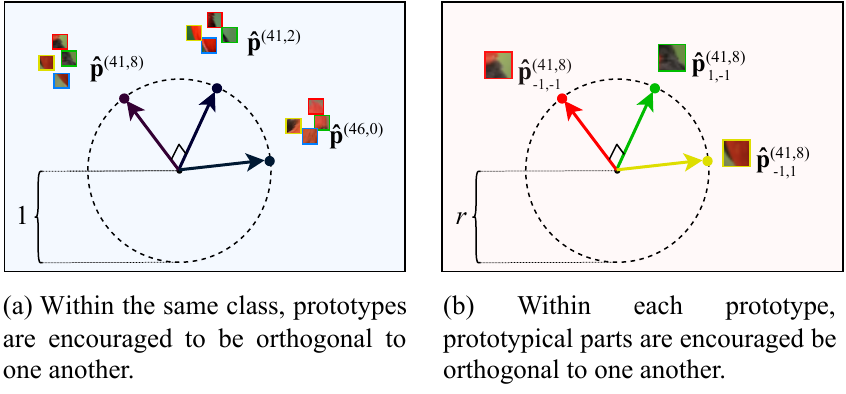}
\captionsetup{width=0.45\textwidth}
  \caption{A representation of the latent space learned by Deformable PrototPNet. 
  }
  \label{fig:latent_space}
\end{figure}

While the subtractive margin encourages separation between classes, it does not encourage diversity between prototypes within a class and between prototypical parts within a prototype. In particular, we have observed that deformations without further regularization often result in duplications of prototypical parts within a prototype. Inspired by \cite{TesNet}, we discourage this behavior by introducing orthogonality loss between prototypical parts. This is formulated as:
\begin{equation}
\label{eq:orthogonality_loss}
    \ell_{\text{ortho}} = \sum_{c} \|\mathbf{P}^{(c)}\mathbf{P}^{(c)\top} -r^2\mathbf{I}^{(\rho L)}\|^2_F ,
\end{equation}
where $L$ is the number of deformable prototypes in class $c$, $\rho L$ is the total number of prototypical parts from all prototypes of class $c$,  $\mathbf{P}^{(c)} \in \mathbb{R}^{\rho L \times d}$
is a matrix with every prototypical part of every prototype from class $c$ arranged as a row in the matrix, and $\mathbf{I}^{(\rho L)}$ is the $\rho L \times \rho L$ identity matrix. The matrix multiplication $\mathbf{P}^{(c)}\mathbf{P}^{(c)\top}$ in equation (\ref{eq:orthogonality_loss}) contains an inner product between every pair of prototypical parts in class $c$; by encouraging this to be close to the scaled identity matrix $r^2\mathbf{I}^{(\rho L)}$, we encourage the prototypical parts to be orthogonal to one another and thereby increase the diversity of semantic concepts represented by prototypical parts. 
This loss differs from \cite{TesNet} because it encourages orthogonality at both the prototype and the prototypical part level. Whereas the orthogonality loss in \cite{TesNet} encourages orthogonality between each pair of prototypes within a class, equation (\ref{eq:orthogonality_loss}) encourages orthogonality between \textit{all prototypical parts} within a class. 
A visualization of the space created by these terms can be seen in Figure \ref{fig:latent_space}.

With these loss terms defined, our overall loss term during the first stage of training is: 
\begin{equation}
\label{eqn:overall_subtractive_loss}
    \ell = \text{CE}^{(-)} + \lambda_1 \ell_{\text{sep}} + \lambda_2 \ell_{\text{clst}} + \lambda_3 \ell_{\text{ortho}}
\end{equation}
where $\lambda_1=0.01, \lambda_2=0.1$ and $\lambda_3=0.1$ are hyperparameters chosen empirically. As in \cite{ProtoPNet}, the last layer connection between each deformable prototype and its class is set to $1$; all other connections are set to $-0.5$.

\textbf{Stage 2: Projection of prototypes.} We project each deformable prototype $\hat{\mathbf{p}}^{(c,l)}$ onto the most similar collection of interpolated image features $\hat{\mathbf{z}}^{(\Delta)}_{a,b}$ from some training image $\mathbf{x}$. Mathematically, this is formulated as:
\begin{equation}
\label{eqn:projection}
    \mathbf{p}^{(c,l)} \leftarrow \underset{\mathbf{\hat{z}}^{(\Delta)}_{a,b}}{\text{argmax}} \  \cos{(\theta(\mathbf{\hat{p}}^{(c,l)} \cdot \mathbf{\hat{z}}^{(\Delta)}_{a,b}))}.
\end{equation}
In this projection scheme, we allow projection onto fractional locations and we project all prototypical parts within each prototype onto the \textit{same} training image, which promotes cohesion among parts of a single prototype.

\textbf{Stage 3: Optimization of the last layer.} In this stage, we fix all other model parameters and optimize over the last layer connections $h$. Let $w^{((c,l), c')}_h$ be defined as previously described. For this stage, we use the loss function:
\begin{equation}
\label{eqn:last_layer_optim}
    \ell_{\text{last}} = \text{CE} + \lambda_1 \sum_{c,l}\sum_{c' \neq c}| w^{((c,l), c')}_h |,
\end{equation}
where $\lambda_1=10^{-3}$ and CE is standard cross entropy loss. The second term on the right-hand side of equation (\ref{eqn:last_layer_optim}) discourages negative reasoning processes as explained in \cite{ProtoPNet}.

\begin{table*}[t]
    \centering
    \begin{tabular}{|p{0.25\textwidth}|c|c|c|c|c|c|c|}
    \hline
    Model & VGG16 & VGG19 & Res34 & Res50 & Res152 & Dense121 & Dense161 \\ 
    \hline
    Baseline & 70.9 & 71.3 & 76.0 &  78.7 &  79.2  & 78.2 & 80.0 \\ \hline 
    ProtoPNet \cite{ProtoPNet} & 70.3* & 72.6* & 72.4* & 81.1* & 74.3* & 74.0* & 75.4* \\\hline 
    Def. ProtoPNet ($3 \times 3$,nd) & 67.9 & 71.1 & 76.7 & 85.9 & 78.2 & 76.5 & 79.6 \\ 
    \hline
    Def. ProtoPNet ($3 \times 3$) & 73.8 & 75.4 & 76.7 & 86.1 & 78.8 & 76.4 & 79.7 \\ \hline
    Def. ProtoPNet ($2 \times 2$,nd)& \textbf{76.0} & \textbf{76.1} & \textbf{76.8} & \textbf{86.4} & 79.2 & 78.9 & 80.8 \\ \hline
    Def. ProtoPNet ($2 \times 2$)& 75.7 & 76.0 & \textbf{76.8} & \textbf{86.4} & \textbf{79.6} & \textbf{79.0} & \textbf{81.2} \\ \hline
    \end{tabular}
    \caption{Accuracy of Deformable ProtoPNets with $3 \times 3$ and $2 \times 2$ deformable prototypes, compared to that of the baseline models, ProtoPNets, and Deformable ProtoPNets without deformations (denoted (nd)) across different base architectures. *We retrained ProtoPNets on full images for direct comparison, and report the accuracy numbers on full images here, so the numbers differ from those reported in \cite{ProtoPNet}.}
    \label{tab:architecture_comparison}
\end{table*}
\subsection{Prototype Visualizations}

With prototype projection, we can associate each deformable prototype $\hat{\mathbf{p}}^{(c,l)}$ with a training image $\mathbf{x}$. Before we describe how we map a prototypical part to an image patch, we first define a downsampling factor $\gamma$ as the ratio of spatial downsampling between the original image and the image-feature tensor. For images of spatial size $224 \times 224$ with latent representations of spatial size of $14 \times 14$, we have $\gamma = \frac{224}{14} = 16$. 

In order to produce a visualization of a deformable prototype on an input image $\mathbf{x}$, we pass the image $\mathbf{x}$ through the network. This enables us to obtain the center location $(a', b')$ that produced the best similarity for prototype $\hat{\mathbf{p}}^{(c,l)}$:
\begin{equation*}
\label{eqn:visualization}
    (a',b') =\underset{a,b}{\text{argmax}}\  g(\mathbf{\hat{z}}^{(\Delta)}_{a,b})^{(c,l)}.
\end{equation*}
We can then retrieve the $(\Delta_1, \Delta_2)$ offset pair for each prototypical part $\hat{\mathbf{p}}^{(c,l)}_{m,n}$ from the location $(a', b')$ of the offset field $\delta(\mathbf{\hat{z}})$. These values tell us that the prototypical part $\hat{\mathbf{p}}^{(c,l)}_{m,n}$ is compared to the image features at spatial location $(a' +m + \Delta_1, b' + n + \Delta_2)$. To find the corresponding patch in the original image, we create a square bounding box in the original image centered at $(\gamma(a'+m+\Delta_1), \gamma (b'+n+\Delta_2))$ of height and width $\gamma$ for each prototypical part $\hat{\mathbf{p}}^{(c,l)}_{m,n}$. Since all parts of a deformable prototype must be projected onto (interpolated) image features from the same image, this allows us to view all parts of a prototype on the same image.

\subsection{Reasoning Process}

\begin{figure}[t]
  \centering
    \includegraphics[width=0.47\textwidth]{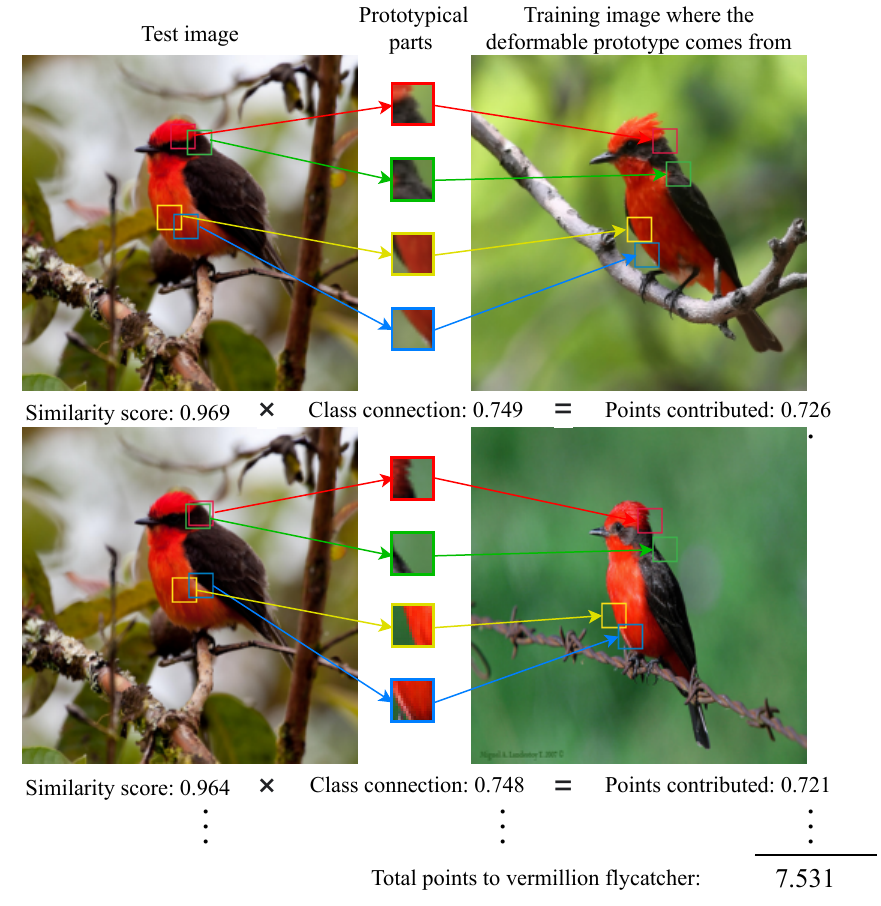}
  \caption{The reasoning process of a Deformable ProtoPNet with $2 \times 2$ deformable prototypes.}
  \label{fig:reasoning_process}
\end{figure}

Figure \ref{fig:reasoning_process} shows the reasoning process of a Deformable ProtoPNet in classifying a test image $\mathbf{x}$. In particular, for a given image $\mathbf{x}$ and for every class $c$, a Deformable ProtoPNet tries to find evidence for $\mathbf{x}$ belonging to class $c$, by comparing the latent features $\hat{\mathbf{z}}$ with every learned deformable prototype $\hat{\mathbf{p}}^{(c, l)}$ of class $c$. In Figure \ref{fig:reasoning_process}, our Deformable ProtoPNet tries to find evidence for the test image being a vermilion flycatcher by comparing the image's latent features with each deformable prototype (whose constituent prototypical parts are visualized in the ``Prototypical parts'' column) of that class. As shown in the figure, the prototypical parts within a deformable prototype, which can be visualized as patches from some training image, can adaptively change their relative spatial positions as the deformable prototype is scanned across the input image to compute a prototype similarity score at each center location according to equation (\ref{eq:deform_similarity}). The maximum score across all spatial locations is taken according to equation (\ref{eq:overall_similarity}), producing a single ``similarity score'' for the prototype, which is multiplied by a class connection score from the fully connected layer $h$ to produce a prototype contribution score. These are summed across all prototypes, yielding a final score for the class.

\section{Experiments and Numerical Results}

We conducted a case study of our Deformable ProtoPNet on the full (uncropped) CUB-200-2011 bird species classification dataset \cite{cub_200}. We trained Deformable ProtoPNets with 6 $3 \times 3$ deformable prototypes per class, and Deformable ProtoPNets with 10 $2 \times 2$ deformable prototypes per class, unless otherwise specified. We ran experiments using VGG \cite{vgg}, ResNet \cite{resnet}, and DenseNet \cite{densenet} as CNN backbones $f$. The ResNet-50 backbone was pretrained on iNaturalist \cite{inaturalist_2018}, and all other backbones were pretrained using ImageNet \cite{deng2009imagenet}. See the supplement for more details regarding our experimental setup.
%

\textbf{We find that Deformable ProtoPNet can achieve competitive accuracy across multiple backbone architectures}. As shown in Table \ref{tab:architecture_comparison}, our Deformable ProtoPNet achieves higher accuracy than ProtoPNet \cite{ProtoPNet} and the non-interpretable baseline model in all cases. For all backbone architectures except VGG-16 and VGG-19 \cite{vgg}, a Deformable ProtoPNet with deformations and $2\times2$ deformable prototypes has the best performance across the models with the same backbone. We ran additional experiments on Stanford Dogs \cite{dogs} and found that our Deformable ProtoPNet also performs well across multiple backbone architectures on that dataset. See the supplement for details.

\begin{table}[t]
    \centering
    \begin{tabular}{|c|c|c||c|}
        \hline
        Margin & Ortho Loss & Deformations & Accuracy\\
        \hline
        \hline
        0.1 & 0 & No & 86.2\\
        \hline
        0.1 & 0 & Yes & \textbf{86.4}\\
        \hline
        0.1 & 0.1 & No & \textbf{86.4}\\
        \hline
        0.1 & 0.1 & Yes & \textbf{86.4}\\
        \hline
        \hline
        0 & 0 & Yes & 86.1\\
        \hline
        0.1 & 0 & Yes & \textbf{86.4}\\
        \hline
        0 & 0.1 & Yes & 85.2\\
        \hline
        0.1 & 0.1 & Yes & \textbf{86.4}\\
        \hline
    \end{tabular}
    \caption{Ablation studies using $2 \times 2$ prototypes.}
    \label{tab:resnet50_ablation_def}
\end{table}

\textbf{We find that using deformations, orthogonality loss, and subtractive margin generally improves (or maintains) accuracy}. As shown in Table \ref{tab:architecture_comparison}, introducing deformations improves (or maintains) accuracy for most backbone architectures. We performed additional ablation studies on ResNet-50-based Deformable ProtoPNets with and without deformations -- these models were trained using various settings of the subtractive margin and orthogonality loss. As shown in Table \ref{tab:resnet50_ablation_def} (top), introducing deformations improves (or maintains) accuracy under the same settings of margin and orthogonality loss. As shown in Table \ref{tab:resnet50_ablation_def} (bottom), introducing subtractive margin generally improves accuracy under the same settings of orthogonality loss and deformations. As shown in both Table \ref{tab:resnet50_ablation_def} (top) and Table \ref{tab:resnet50_ablation_def} (bottom), introducing orthogonality loss maintains accuracy in most cases.

\textbf{We find that Deformable ProtoPNet can achieve state-of-the-art accuracy.} As Table \ref{tab:state-of-the-art} (top) shows, a single Deformable ProtoPNet can achieve high accuracy ($86.1\%$ with 6 $3 \times 3$ prototypes per class, $86.4\%$ with 10 $2 \times 2$ prototypes per class) on full test images from CUB-200-2011 \cite{cub_200}, outperforming a single ProtoTree \cite{ProtoTree} ($82.2\%$) and 3 ensembled TesNets \cite{TesNet} ($83.5\%$). Additionally, 5 ensembled Deformable ProtoPNets using $2 \times 2$ prototypes outperform all competing models, achieving state-of-the-art accuracy ($87.8\%$). Table \ref{tab:state-of-the-art} (bottom) shows that Deformable ProtoPNet also performs well on Stanford Dogs \cite{dogs}, achieving accuracy ($86.5\%$) competitive with the state-of-the-art.


\begin{table}[t]
    \centering
    \begin{tabular}{|l|l|}
    \hline
    Interpretability & Model: accuracy on CUB-200-2011                  
    
    \\ \hline
    None & B-CNN\cite{lin2015BCNN}: 85.1 (bb), 84.1 (f)                                         
    \\ \hline
    \begin{tabular}[c]{@{}l@{}}Object-level \\attention\end{tabular}
    
    & \begin{tabular}[c]{@{}l@{}}
     CAM\cite{zhou2016CAM}: 70.5 (bb), 63.0 (f)\\        
     \textbf{CSG}\cite{liang2020CSG} \textbf{82.6 (bb), \textbf{78.5} (f)} 
    \end{tabular} \\ \hline
    
    \begin{tabular}[c]{@{}l@{}}Part-level\\ attention \end{tabular}                                                                            & \begin{tabular}[c]{@{}l@{}} PA-CNN\cite{krause2015PACNN}: 82.8 (bb)\\ 
    MG-CNN\cite{wang2015MGCNN}: 83.0 (bb), 81.7 (f)\\
    MA-CNN\cite{zheng2017MACNN}: 86.5 (f)\\
    RA-CNN\cite{fu2017RACNN}: 85.3  (f)\\ \textbf{TASN}\cite{zheng2019TASN}: \textbf{87.0 (f)}\end{tabular} \\ \hline
    
    \begin{tabular}[c]{@{}l@{}}Part-level \\attention +\\  prototypes\end{tabular} &
    
    \begin{tabular}[c]{@{}l@{}} 
        Region\cite{huang2020region}: 81.5 (bb), 80.2 (f)\\
        ProtoPNet*\cite{ProtoPNet}: 84.8 (bb), 81.1 (f)\\
        ProtoTree\cite{ProtoTree}: 82.2 (f)\\
        \textbf{ProtoTree**\cite{ProtoTree}}: \textbf{87.2 (f)}\\
        TesNet*\cite{TesNet}: 86.2 (bb), 83.5 (f)\\
        Def. ProtoPNet [nd,3p,6pc]: 85.9 (f)\\
        Def. ProtoPNet [nd,2p,10pc]: 86.4 (f)
    \end{tabular} \\
    \hline
    
    \begin{tabular}[c]{@{}l@{}}Part-level \\attention. +\\  prototypes +\\ deformations\end{tabular} &
    
    \begin{tabular}[c]{@{}l@{}} 
        Def. ProtoPNet [3p,1pc]: 81.5 (f)\\
        Def. ProtoPNet [3p,3pc]: 83.7 (f)\\
        Def. ProtoPNet [3p,6pc]: 86.1 (f)\\
        Def. ProtoPNet [2p,10pc]: 86.4 (f)\\
        \textbf{Def. ProtoPNet** [2p,10pc]}: \textbf{87.8 (f)}
    \end{tabular}
    
    
    
    \\ \hline
    \hline
    Interpretability & Model: accuracy on Stanford Dogs             \\ \hline
    
    \begin{tabular}[c]{@{}l@{}}Part-level\\ attention \end{tabular}                                                                            & \begin{tabular}[c]{@{}l@{}} FCAN\cite{FCAN}: 84.2\\ 
    \textbf{RA-CNN}\cite{fu2017RACNN}: \textbf{87.3}\end{tabular} \\ \hline
    
    \begin{tabular}[c]{@{}l@{}}Part-level \\attention +\\  prototypes\end{tabular} &
    
    \begin{tabular}[c]{@{}l@{}} 
        ProtoPNet\cite{ProtoPNet}: 77.3\\
        \textbf{Def. ProtoPNet[nd,3p,10pc]}: \textbf{86.5}
    \end{tabular} \\
    \hline
    
    \begin{tabular}[c]{@{}l@{}}Part-level \\attention +\\  prototypes +\\ deformations\end{tabular} &
    
    \begin{tabular}[c]{@{}l@{}} 
        \textbf{Def. ProtoPNet[3p,10pc]}: \textbf{86.5}
    \end{tabular}
    
    
    
    \\ \hline
    \end{tabular}
    \caption{Accuracy and interpretability of Deformable ProtoPNet compared to other models on CUB-200-2011 (top) and Stanford Dogs (bottom). Methods using bounding boxes are marked (bb) and methods using full, uncropped images are marked (f). For the Deformable ProtoPNets, we denote $k$ prototypes per class as $k$pc, Deformable ProtoPNets with $2 \times 2$ protoypes as 2p, Deformable ProtoPNets with $3 \times 3$ protoypes as 3p, and Deformable ProtoPNets without deformations as nd.
    *Using 3 ensembled models. **Using 5 ensembled models.}
    \label{tab:state-of-the-art}
\end{table}


\section{Conclusion}
We presented Deformable ProtoPNet, a case-based interpretable neural network with deformable prototypes. The competitive performance and transparency of this model will enable wider use of interpretable models for computer vision. One limitation of Deformable ProtoPNet is that offsets are shared across all deformable prototypes at each spatial location. Another limitation is that we observed semantic mismatches between some prototypical parts and image parts that are considered ``similar'' by Deformable ProtoPNet. We plan to address these limitations in future work.

\textbf{Acknowledgment.} This work was supported in part by the GPU cluster, funded by NSF MRI \# 1919478, at the Advanced Computing Group at the University of Maine.

\FloatBarrier
{\small
\bibliographystyle{ieee_fullname}
\bibliography{references}
}

\section*{Supplementary Material}

%
\crefname{section}{Sec.}{Secs.}
\Crefname{section}{Section}{Sections}
\Crefname{table}{Table}{Tables}
\crefname{table}{Tab.}{Tabs.}

\section{Proof of Theorem 3.1}


\textbf{Theorem 3.1} Let $\mathbf{\hat{z}}_1, \mathbf{\hat{z}}_2, \mathbf{\hat{z}}_3, \mathbf{\hat{z}}_4 \in \mathbb{R}^n$ be vectors such that $\|\mathbf{\hat{z}}_i\| = r$ for all $i \in \{1,2,3,4\}$ for some constant $r$, and let $\hat{\mathbf{z}}^2$ denote the element-wise square of a vector. For some constants $\alpha \in [0, 1]$ and $\beta \in [0,1]$, the bilinear interpolation operation $\mathbf{z}_{\text{interp}}=(1-\alpha)(1-\beta)\mathbf{\hat{z}}_1 + (1 - \alpha)\beta\mathbf{\hat{z}}_2 + \alpha(1-\beta)\mathbf{\hat{z}}_3 + \alpha\beta\mathbf{\hat{z}}_4$ does not guarantee that $\|\mathbf{z}_{\text{interp}}\|_2 = r$. However, the $L^2$ norm-preserving interpolation operation $\mathbf{z}_{\text{interp}}=\sqrt{(1-\alpha)(1-\beta)\mathbf{\hat{z}}^2_1 + (1 - \alpha)\beta\mathbf{\hat{z}}^2_2 + \alpha(1-\beta)\mathbf{\hat{z}}^2_3 + \alpha\beta\mathbf{\hat{z}}^2_4}$ guarantees that $\|\mathbf{z}_{\text{interp}}\|_2 = r$ for all $\alpha \in [0, 1]$ and $\beta \in [0,1]$. (The square root is taken element-wise.)

\begin{proof}
To show that bilinear interpolation of vectors with the same $L^2$ norm does \textit{not} in general preserve the $L^2$ norm, consider the following example with:
\[
\mathbf{\hat{z}}_1 = \begin{bmatrix} & r \\ & 0 \\ & 0 \\ & 0 \end{bmatrix}, \mathbf{\hat{z}}_2 = \begin{bmatrix} & 0 \\ & r \\ & 0 \\ & 0 \end{bmatrix}, \mathbf{\hat{z}}_3 = \begin{bmatrix} & 0 \\ & 0 \\ & r \\ & 0 \end{bmatrix}, \mathbf{\hat{z}}_4 = \begin{bmatrix} & 0 \\ & 0 \\ & 0 \\ & r \end{bmatrix}.
\]
For $\alpha = \beta = 1/2$, using bilinear interpolation, we have:
\begin{align*}
\|\mathbf{z}_{\text{interp}}\|_2^2 &= \|(1-\alpha)(1-\beta)\mathbf{\hat{z}}_1 + (1 - \alpha)\beta\mathbf{\hat{z}}_2 \\ &\indent\indent + \alpha(1-\beta)\mathbf{\hat{z}}_3 + \alpha\beta\mathbf{\hat{z}}_4\|_2^2 \\
&= \left\lVert \frac{1}{4}\begin{bmatrix} & r \\ & 0 \\ & 0 \\ & 0 \end{bmatrix} + \frac{1}{4}\begin{bmatrix} & 0 \\ & r \\ & 0 \\ & 0 \end{bmatrix} + \frac{1}{4}\begin{bmatrix} & 0 \\ & 0 \\ & r \\ & 0 \end{bmatrix} + \frac{1}{4}\begin{bmatrix} & 0 \\ & 0 \\ & 0 \\ & r \end{bmatrix}\right\rVert_2^2 \\
&= \left\lVert \begin{bmatrix} & \frac{1}{4}r \\ & \frac{1}{4}r \\ & \frac{1}{4}r \\ & \frac{1}{4}r \end{bmatrix} \right\rVert_2^2 \\
&= \left(\frac{1}{4}r\right)^2 + \left(\frac{1}{4}r\right)^2 + \left(\frac{1}{4}r\right)^2 + \left(\frac{1}{4}r\right)^2 \\
&= \frac{1}{4}r^2,
\end{align*}
which means 
\[
\|\mathbf{z}_{\text{interp}}\|_2 = \sqrt{\frac{1}{4}r^2} = \frac{1}{2}r \neq r.
\]
This counter-example shows that bilinear interpolation generally does not preserve the $L^2$ norm of the interpolated vector.

Now, we show that the $L^2$ norm \hbox{preserving} interpolation operation $\mathbf{z}_{\text{interp}}=\sqrt{(1-\alpha)(1-\beta)\mathbf{\hat{z}}^2_1 + (1 - \alpha)\beta\mathbf{\hat{z}}^2_2 + \alpha(1-\beta)\mathbf{\hat{z}}^2_3 + \alpha\beta\mathbf{\hat{z}}^2_4}$, when applied to vectors $\mathbf{\hat{z}}_1, \mathbf{\hat{z}}_2, \mathbf{\hat{z}}_3, \mathbf{\hat{z}}_4 \in \mathbb{R}^n$ with the same $L^2$ norm $r$, yields an interpolated vector of the same $L^2$ norm $r$. Let
\[
\mathbf{\hat{z}}_i = \begin{bmatrix} & \hat{z}_{i,1} \\ & \hat{z}_{i,2} \\ & \vdots \\ & \hat{z}_{i,n} \end{bmatrix} \text{ for all } i \in \{1, 2, 3, 4\},
\]
where $\hat{z}_{i,j}$ denotes the $j$-th component of the vector $\mathbf{\hat{z}}_i$. The element-wise square of the vector $\mathbf{\hat{z}}_i$ is given by:
\[
\mathbf{\hat{z}}_i^2 = \begin{bmatrix} & \hat{z}_{i,1}^2 \\ & \hat{z}_{i,2}^2 \\ & \vdots \\ & \hat{z}_{i,n}^2 \end{bmatrix} \text{ for all } i \in \{1, 2, 3, 4\}.
\]
Using the $L^2$ norm preserving interpolation on $\mathbf{\hat{z}}_1, \mathbf{\hat{z}}_2, \mathbf{\hat{z}}_3, \mathbf{\hat{z}}_4 \in \mathbb{R}^n$ with $\|\mathbf{\hat{z}}_i\| = r$ for all $i \in \{1,2,3,4\}$, \textit{for all} $\alpha \in [0, 1]$ and $\beta \in [0, 1]$, and letting $\tilde{\alpha} = 1 - \alpha$ and $\tilde{\beta} = 1 - \beta$, we have:
\begin{align*}
&\indent \|\mathbf{z}_{\text{interp}}\|_2^2 \\
&= \left\lVert \sqrt{\tilde{\alpha}\tilde{\beta}\mathbf{\hat{z}}^2_1 + \tilde{\alpha}\beta\mathbf{\hat{z}}^2_2 + \alpha\tilde{\beta}\mathbf{\hat{z}}^2_3 + \alpha\beta\mathbf{\hat{z}}^2_4}\right\rVert_2^2 \\
&= \left\lVert \begin{bmatrix} & \sqrt{\tilde{\alpha}\tilde{\beta}\hat{z}_{1,1}^2 + \tilde{\alpha}\beta\hat{z}_{2,1}^2 + \alpha\tilde{\beta}\hat{z}_{3,1}^2 + \alpha\beta\hat{z}_{4,1}^2} \\ & \sqrt{\tilde{\alpha}\tilde{\beta}\hat{z}_{1,2}^2 + \tilde{\alpha}\beta\hat{z}_{2,2}^2 + \alpha\tilde{\beta}\hat{z}_{3,2}^2 + \alpha\beta\hat{z}_{4,2}^2} \\ & \vdots \\ & \sqrt{\tilde{\alpha}\tilde{\beta}\hat{z}_{1,n}^2 + \tilde{\alpha}\beta\hat{z}_{2,n}^2 + \alpha\tilde{\beta}\hat{z}_{3,n}^2 + \alpha\beta\hat{z}_{4,n}^2} \end{bmatrix}\right\rVert_2^2 \\
&= \indent \tilde{\alpha}\tilde{\beta}\hat{z}_{1,1}^2 + \tilde{\alpha}\beta\hat{z}_{2,1}^2 + \alpha\tilde{\beta}\hat{z}_{3,1}^2 + \alpha\beta\hat{z}_{4,1}^2 \\
&\indent + \tilde{\alpha}\tilde{\beta}\hat{z}_{1,2}^2 + \tilde{\alpha}\beta\hat{z}_{2,2}^2 + \alpha\tilde{\beta}\hat{z}_{3,2}^2 + \alpha\beta\hat{z}_{4,2}^2  \\
&\indent ...\\
&\indent + \tilde{\alpha}\tilde{\beta}\hat{z}_{1,n}^2 + \tilde{\alpha}\beta\hat{z}_{2,n}^2 + \alpha\tilde{\beta}\hat{z}_{3,n}^2 + \alpha\beta\hat{z}_{4,n}^2 \\
&= \tilde{\alpha}\tilde{\beta}\|\mathbf{\hat{z}}_1\|_2^2  + \tilde{\alpha}\beta\|\mathbf{\hat{z}}_2\|_2^2 + \alpha\tilde{\beta}\|\mathbf{\hat{z}}_3\|_2^2 + \alpha\beta\|\mathbf{\hat{z}}_4\|_2^2 \\
&= (1-\alpha)(1-\beta)r^2 + (1-\alpha)\beta r^2 + \alpha(1-\beta)r^2 + \alpha\beta r^2 \\
&= (1-\alpha)r^2 + \alpha r^2 \\
&= r^2,
\end{align*}
which means
\[
\|\mathbf{z}_{\text{interp}}\|_2 = \sqrt{r^2} = r,
\]
thus completing the proof.
\end{proof}

In a Deformable ProtoPNet, we apply the $L^2$ norm preserving interpolation operation to compute the interpolated image features $\mathbf{\hat{z}}_{a+m+\Delta_1, b+n+\Delta_2}$ for given values of $a$, $b$, $m$, and $n$, as follows:
\begin{align*}
&\indent \mathbf{\hat{z}}_{a+m+\Delta_1, b+n+\Delta_2} \\
&= \sqrt{(1-\alpha)(1-\beta)\mathbf{\hat{z}}^2_1 + (1 - \alpha)\beta\mathbf{\hat{z}}^2_2 + \alpha(1-\beta)\mathbf{\hat{z}}^2_3 + \alpha\beta\mathbf{\hat{z}}^2_4}
\end{align*}
with
\begin{equation}\label{eq:supp-z1}
\mathbf{\hat{z}}_1 = \mathbf{\hat{z}}_{\lfloor a+m+\Delta_1 \rfloor, \lfloor b+n+\Delta_2 \rfloor},
\end{equation}
\begin{equation}\label{eq:supp-z2}
\mathbf{\hat{z}}_2 = \mathbf{\hat{z}}_{\lfloor a+m+\Delta_1 \rfloor, \lceil b+n+\Delta_2 \rceil},
\end{equation}
\begin{equation}\label{eq:supp-z3}
\mathbf{\hat{z}}_3 = \mathbf{\hat{z}}_{\lceil a+m+\Delta_1 \rceil, \lfloor b+n+\Delta_2 \rfloor},
\end{equation}
\begin{equation}\label{eq:supp-z4}
\mathbf{\hat{z}}_4 = \mathbf{\hat{z}}_{\lceil a+m+\Delta_1 \rceil, \lceil b+n+\Delta_2 \rceil},
\end{equation}
and
\begin{equation}\label{eq:supp-alpha}
\alpha = (a+m+\Delta_1) - \lfloor a+m+\Delta_1 \rfloor = \Delta_1 - \lfloor \Delta_1 \rfloor,
\end{equation}
\begin{equation}\label{eq:supp-beta}
\beta = (b+n+\Delta_2) - \lfloor b+n+\Delta_2 \rfloor = \Delta_2 - \lfloor \Delta_2 \rfloor.
\end{equation}

\section{Backpropagation through a Deformable Prototype}


Recall that a deformable prototype $\mathbf{\hat{p}}^{(c,l)}$, when applied at the spatial position $(a, b)$ on the image-feature tensor $\mathbf{\hat{z}}$, computes its similarity with the interpolated image features $\mathbf{\hat{z}}^{\Delta}_{a,b}$ according to the following equation:
\begin{equation}\label{eq:supp-sim}
g(\mathbf{\hat{z}})^{(c,l)}_{a,b} = \sum_m \sum_n \mathbf{\hat{p}}^{(c,l)}_{m,n} \cdot \mathbf{\hat{z}}_{a+m+\Delta_1, b+n+\Delta_2},
\end{equation}
where we have
\begin{equation}\label{eq:supp-z-interp}
\mathbf{\hat{z}}_{a+m+\Delta_1, b+n+\Delta_2} = \sqrt{\zeta^{(a,b,m,n)}(\mathbf{\hat{z}}, \Delta_1, \Delta_2)}
\end{equation}
and we have defined
\begin{align*}
&\indent \zeta^{(a,b,m,n)}(\mathbf{\hat{z}}, \Delta_1, \Delta_2) \\
&= (1-\alpha)(1-\beta)\mathbf{\hat{z}}^2_1 + (1 - \alpha)\beta\mathbf{\hat{z}}^2_2 + \alpha(1-\beta)\mathbf{\hat{z}}^2_3 + \alpha\beta\mathbf{\hat{z}}^2_4,
\end{align*}
where $\mathbf{\hat{z}}_1$, $\mathbf{\hat{z}}_2$, $\mathbf{\hat{z}}_3$, $\mathbf{\hat{z}}_4$, $\alpha$, and $\beta$ are given by equations (\ref{eq:supp-z1}), (\ref{eq:supp-z2}), (\ref{eq:supp-z3}), (\ref{eq:supp-z4}), (\ref{eq:supp-alpha}), and (\ref{eq:supp-beta}), respectively. Note that $\zeta^{(a,b,m,n)}(\mathbf{\hat{z}}, \Delta_1, \Delta_2)$ can be rewritten as:
\begin{equation}\label{eq:supp-zeta}
\begin{split}
&\indent \zeta^{(a,b,m,n)}(\mathbf{\hat{z}}, \Delta_1, \Delta_2) \\
&= \sum_i \sum_j \mathbf{\hat{z}}_{i,j}^2 \max(0, 1 - |(a+m+\Delta_1) - i|) \\ 
&\indent\indent\indent\indent \cdot \max(0, 1 - |(b+n+\Delta_2) - j|).
\end{split}
\end{equation}


To show that we can back-propagate gradients through a deformable prototype, it is sufficient to show that we can compute the gradients of the prototype similarity score $g(\mathbf{\hat{z}})^{(c,l)}_{a,b}$ (when a deformable prototype $\mathbf{\hat{p}}^{(c,l)}$ is applied at the spatial position $(a, b)$ on the image-feature tensor $\mathbf{\hat{z}}$), with respect to every $(m, n)$-th prototypical part $\mathbf{\hat{p}}^{(c,l)}_{m,n}$ of the deformable prototype $\mathbf{\hat{p}}^{(c,l)}$ and with respect to every (discrete) spatial position $\mathbf{\hat{z}}_{i,j}$ of the image-feature tensor $\mathbf{\hat{z}}$.

From equation (\ref{eq:supp-sim}), it is easy to see that the gradient of the prototype similarity score $g(\mathbf{\hat{z}})^{(c,l)}_{a,b}$ with respect to the $(m, n)$-th prototypical part $\mathbf{\hat{p}}^{(c,l)}_{m,n}$ is given by:
\[
\frac{\partial g(\mathbf{\hat{z}})^{(c,l)}_{a,b}}{\partial \mathbf{\hat{p}}^{(c,l)}_{m,n}} = \mathbf{\hat{z}}_{a+m+\Delta_1, b+n+\Delta_2}^{\top}.
\]

Before we derive the gradient of the prototype similarity score $g(\mathbf{\hat{z}})^{(c,l)}_{a,b}$ with respect to a (discrete) spatial position $\mathbf{\hat{z}}_{i,j}$ of the image-feature tensor $\mathbf{\hat{z}}$, note that the prototype similarity score $g(\mathbf{\hat{z}})^{(c,l)}_{a,b}$ is computed by first computing $\zeta^{(a,b,m,n)}(\mathbf{\hat{z}}, \Delta_1, \Delta_2)$ using equation (\ref{eq:supp-zeta}), followed by computing $\mathbf{\hat{z}}_{a+m+\Delta_1, b+n+\Delta_2}$ by taking the square root of $\zeta^{(a,b,m,n)}(\mathbf{\hat{z}}, \Delta_1, \Delta_2)$ element-wise (equation (\ref{eq:supp-z-interp}), and finally computing the similarity score using equation \ref{eq:supp-sim}. In particular, in the first step when we compute $\zeta^{(a,b,m,n)}(\mathbf{\hat{z}}, \Delta_1, \Delta_2)$ using equation (\ref{eq:supp-zeta}), note that $\Delta_1 = \Delta_1(\mathbf{\hat{z}}, a, b, m, n)$ and $\Delta_2 = \Delta_2(\mathbf{\hat{z}}, a, b, m, n)$ are functions depending on $\mathbf{\hat{z}}$, $a$, $b$, $m$, and $n$ -- in particular, for all (discrete) spatial positions $(a, b)$ on the image-feature tensor $\mathbf{\hat{z}}$ and for all $(m, n)$-th prototypical parts, $\Delta_1$ and $\Delta_2$ are produced by applying convolutional layers to $\mathbf{\hat{z}}$. Hence, there are three ways in which $\mathbf{\hat{z}}$ can influence $\zeta^{(a,b,m,n)}(\mathbf{\hat{z}}, \Delta_1, \Delta_2)$: (1) directly through the $\mathbf{\hat{z}}_{i,j}^2$ term in equation (\ref{eq:supp-zeta}), (2) through $\Delta_1$, and (3) through $\Delta_2$. We have to take into account all three ways in which $\mathbf{\hat{z}}$ influences $\zeta^{(a,b,m,n)}(\mathbf{\hat{z}}, \Delta_1, \Delta_2)$, when applying the chain rule to compute the gradient of the prototype similarity score $g(\mathbf{\hat{z}})^{(c,l)}_{a,b}$ with respect to a (discrete) spatial position $\mathbf{\hat{z}}_{i,j}$ of the image-feature tensor $\mathbf{\hat{z}}$. In particular, we have:
\begin{equation}\label{eq:supp-gradient-zij}
\begin{split}
&\indent \frac{\partial g(\mathbf{\hat{z}})^{(c,l)}_{a,b}}{\partial \mathbf{\hat{z}}_{i,j}} \\
&= \sum_m \sum_n \frac{\partial g(\mathbf{\hat{z}})^{(c,l)}_{a,b}}{\partial \mathbf{\hat{z}}_{a+m+\Delta_1, b+n+\Delta_2}}\frac{\partial \mathbf{\hat{z}}_{a+m+\Delta_1, b+n+\Delta_2}}{\partial\zeta^{(a,b,m,n)}(\mathbf{\hat{z}}, \Delta_1, \Delta_2)}\\
&\indent \Bigg{(}\frac{\partial \zeta^{(a,b,m,n)}(\mathbf{\hat{z}}, \Delta_1, \Delta_2)}{\partial \mathbf{\hat{z}}_{i,j}} \\ &\indent\indent + \frac{\partial \zeta^{(a,b,m,n)}(\mathbf{\hat{z}}, \Delta_1, \Delta_2)}{\partial \Delta_1}\frac{\partial \Delta_1}{\partial \mathbf{\hat{z}}_{i,j}} \\
&\indent\indent\indent + \frac{\partial \zeta^{(a,b,m,n)}(\mathbf{\hat{z}}, \Delta_1, \Delta_2)}{\partial \Delta_2}\frac{\partial \Delta_2}{\partial \mathbf{\hat{z}}_{i,j}}\Bigg{)}.
\end{split}
\end{equation}

From equation (\ref{eq:supp-sim}), for given values of $a$, $b$, $m$, and $n$, we have:
\begin{equation}\label{eq:supp-dg-dzinterp}
\frac{\partial g(\mathbf{\hat{z}})^{(c,l)}_{a,b}}{\partial \mathbf{\hat{z}}_{a+m+\Delta_1, b+n+\Delta_2}} = (\mathbf{\hat{p}}^{(c,l)}_{m,n})^{\top}.
\end{equation}

From equation (\ref{eq:supp-z-interp}), for given values of $a$, $b$, $m$, $n$, we have:
\begin{equation}\label{eq:supp-dzinterp-dzeta}
\frac{\partial \mathbf{\hat{z}}_{a+m+\Delta_1, b+n+\Delta_2}}{\partial \zeta^{(a,b,m,n)}(\mathbf{\hat{z}}, \Delta_1, \Delta_2)} = \text{diag}\Bigg{(}\frac{1}{2\sqrt{\zeta^{(a,b,m,n)}(\mathbf{\hat{z}}, \Delta_1, \Delta_2)}}\Bigg{)},
\end{equation}
where diag is a function that converts a $d$-dimensional vector into a $d \times d$ diagonal matrix, and the square root is taken element-wise.

From equation (\ref{eq:supp-zeta}), for given values of $a$, $b$, $m$, $n$, we have:
\begin{equation}\label{eq:supp-dzeta-dzij}
\begin{split}
&\indent \frac{\partial \zeta^{(a,b,m,n)}(\mathbf{\hat{z}}, \Delta_1, \Delta_2)}{\partial \mathbf{\hat{z}}_{i,j}} \\
&= \text{diag}\big{(}2\mathbf{\hat{z}}_{i,j}\max(0, 1 - |(a+m+\Delta_1) - i|) \\ 
&\indent\indent\indent\indent \cdot \max(0, 1 - |(b+n+\Delta_2) - j|)\big{)},
\end{split}
\end{equation}
\begin{equation}\label{eq:supp-dzeta-dDelta1}
\begin{split}
&\indent \frac{\partial \zeta^{(a,b,m,n)}(\mathbf{\hat{z}}, \Delta_1, \Delta_2)}{\partial \Delta_1} \\
&= \sum_i \sum_j \mathbf{\hat{z}}_{i,j}^2 \max(0, 1 - |(b+n+\Delta_2) - j|) \\
&\indent\indent\indent\indent \cdot \begin{cases}
0  &\text{if }  |(a+m+\Delta_1) - i| \geq 1\\
1  &\text{if } -1 < (a+m+\Delta_1) - i < 0 \\
-1 &\text{if } 0 \leq (a+m+\Delta_1) - i < 1,
\end{cases}
\end{split}
\end{equation}
and
\begin{equation}\label{eq:supp-dzeta-dDelta2}
\begin{split}
&\indent \frac{\partial \zeta^{(a,b,m,n)}(\mathbf{\hat{z}}, \Delta_1, \Delta_2)}{\partial \Delta_2} \\
&= \sum_i \sum_j \mathbf{\hat{z}}_{i,j}^2 \max(0, 1 - |(a+m+\Delta_1) - i|) \\
&\indent\indent\indent\indent \cdot \begin{cases}
0  &\text{if }  |(b+n+\Delta_2) - j| \geq 1\\
1  &\text{if } -1 < (b+n+\Delta_2) - j < 0 \\
-1 &\text{if } 0 \leq (b+n+\Delta_2) - j < 1.
\end{cases}
\end{split}
\end{equation}

With equations (\ref{eq:supp-dg-dzinterp}) -- (\ref{eq:supp-dzeta-dDelta2}), and noting that the gradients $\frac{\partial \Delta_1}{\partial \mathbf{\hat{z}}_{i,j}}$ and $\frac{\partial \Delta_2}{\partial \mathbf{\hat{z}}_{i,j}}$ are well-defined (because $\Delta_1$ and $\Delta_2$ are produced by applying convolutional layers to $\mathbf{\hat{z}}$, and convolutional layers are differentiable), we have shown that the gradient of the prototype similarity score $g(\mathbf{\hat{z}})^{(c,l)}_{a,b}$ with respect to a (discrete) spatial position $\mathbf{\hat{z}}_{i,j}$ of the image-feature tensor $\mathbf{\hat{z}}$, given in equation (\ref{eq:supp-gradient-zij}), is well-defined. Hence, we can back-propagate through a deformable prototype.

\section{More Examples of Reasoning Processes}
\label{sec:supp:reasoning}

\begin{figure}[t]
  \centering
    \includegraphics[width=0.47\textwidth]{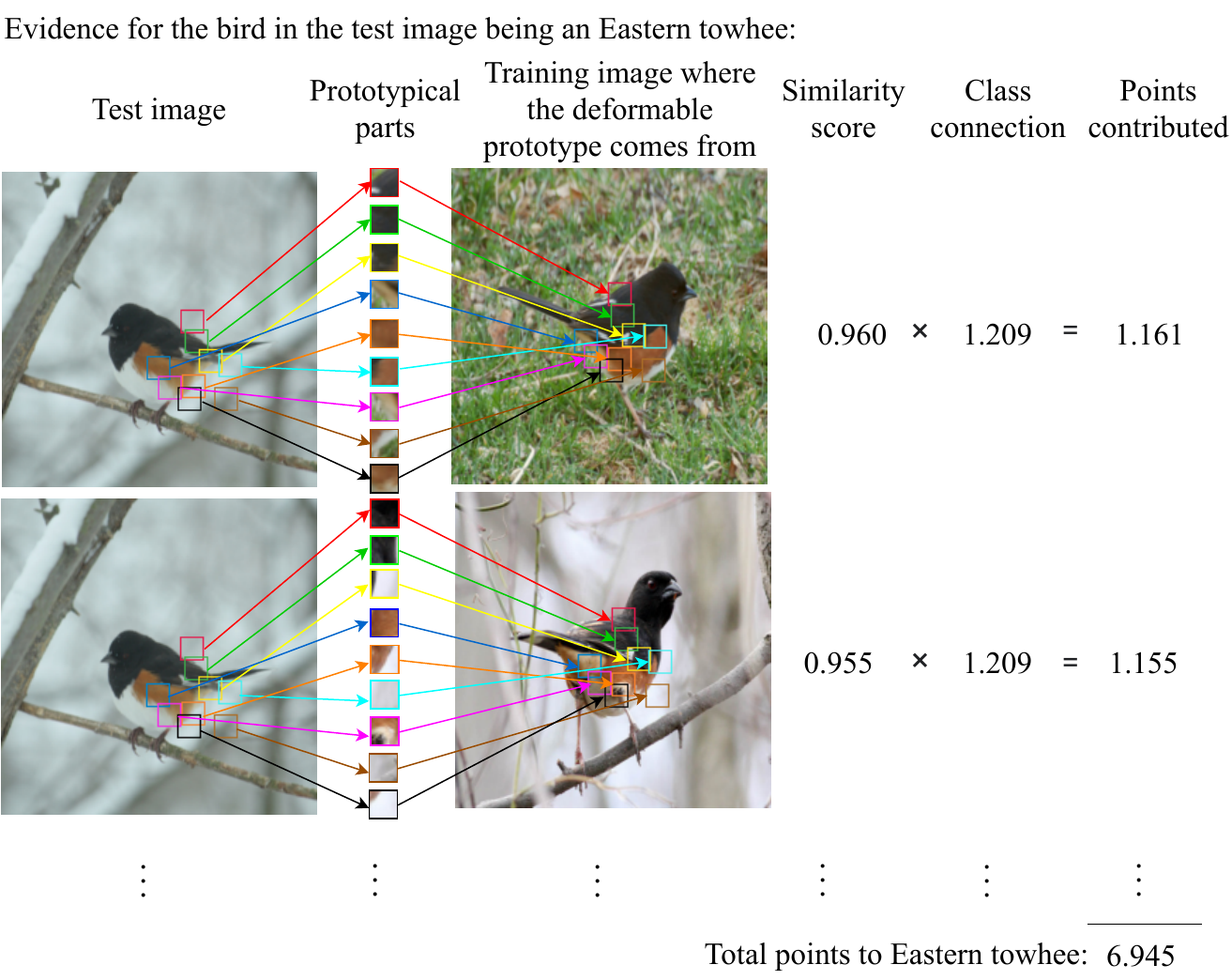}
    \includegraphics[width=0.47\textwidth]{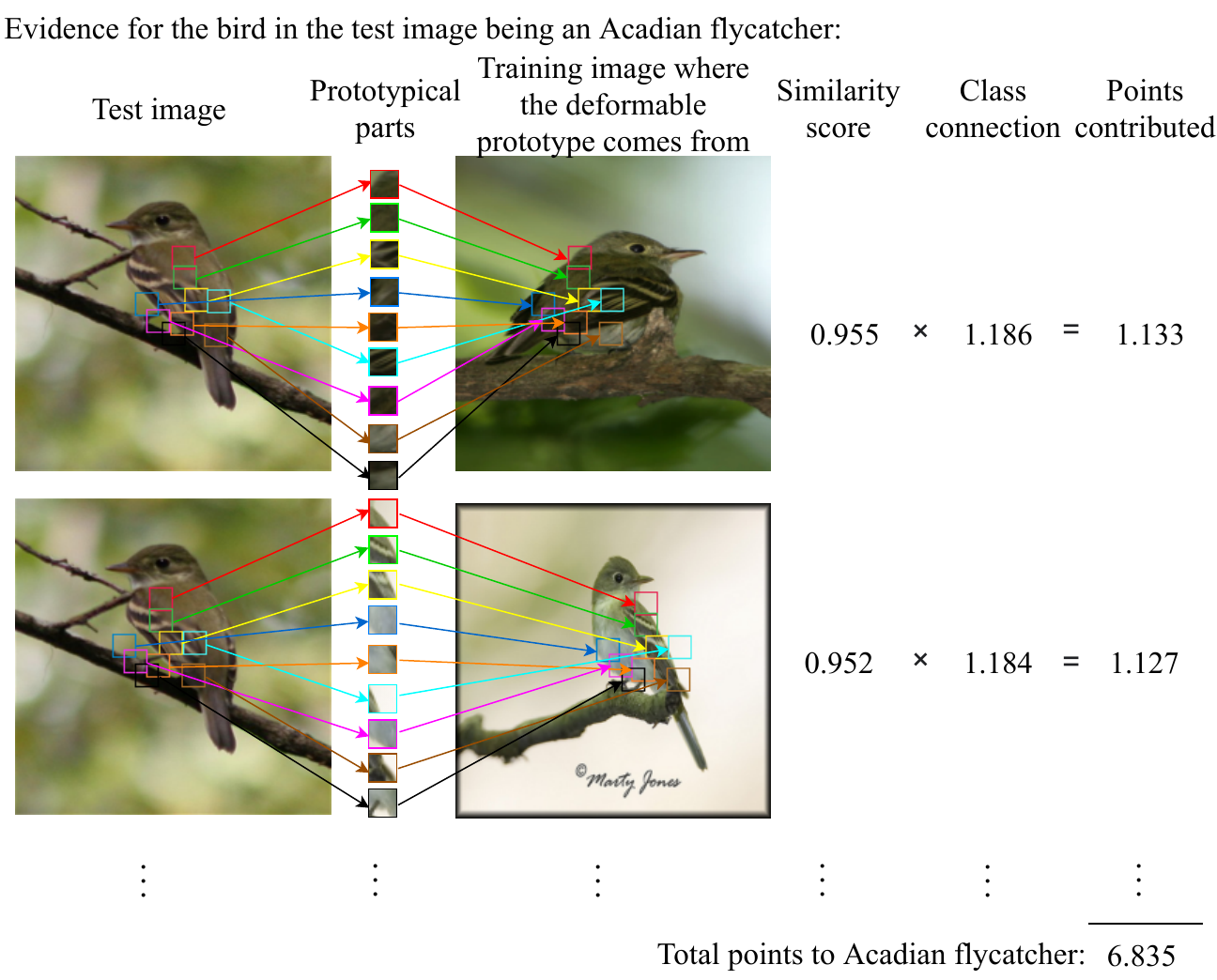}
    \includegraphics[width=0.47\textwidth]{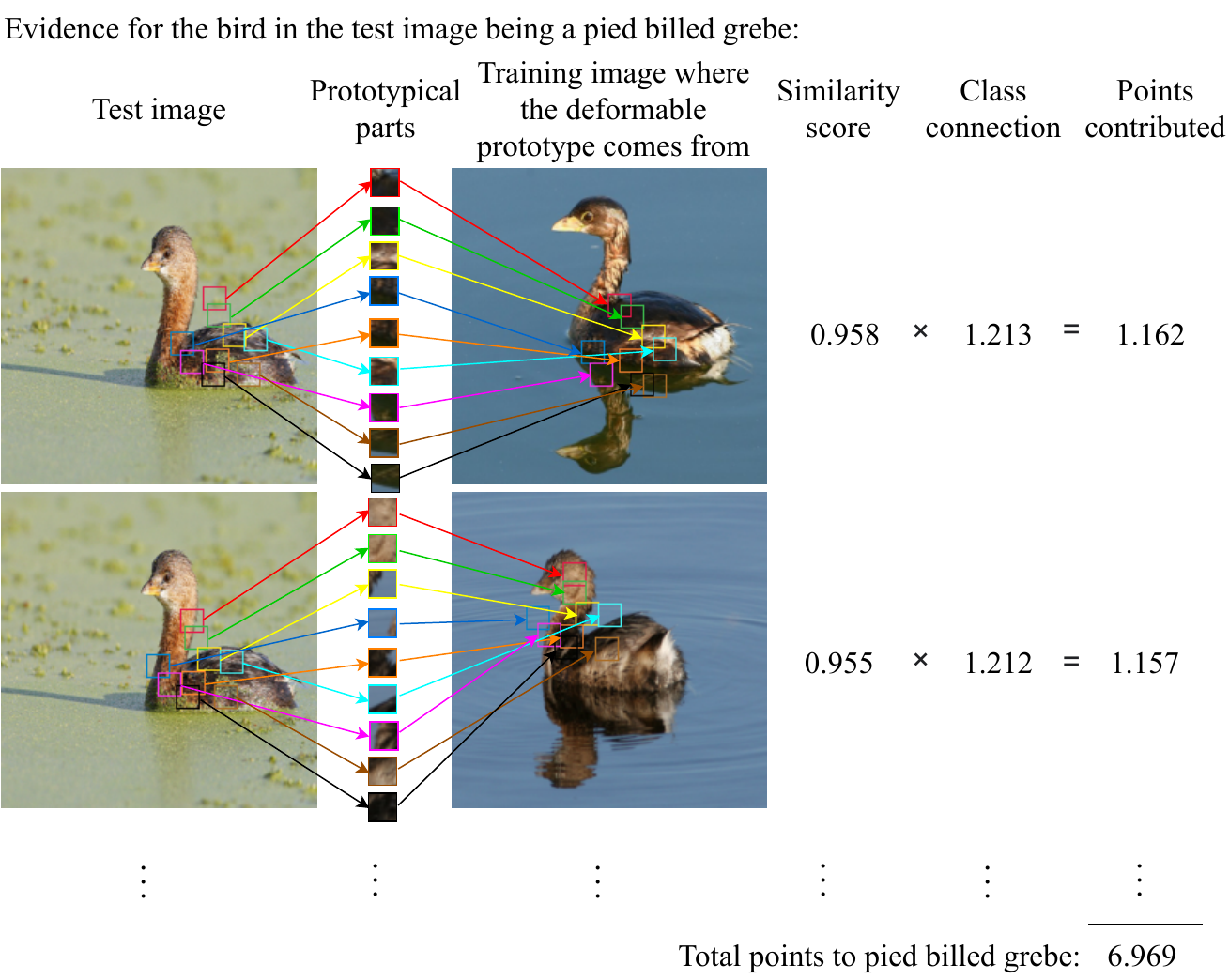}
  \caption{Example reasoning processes of a Deformable ProtoPNet with $3 \times 3$ prototypes when classifying a test image of an Eastern towhee (top), an Acadian flycatcher (middle), and a pied billed grebe (bottom). In each case, we show the two deformable prototypes of the predicted class that produced the highest similarity scores.}
  \label{fig:reasoning_process_birds}
\end{figure}

\begin{figure}[t]
  \centering
    \includegraphics[width=0.47\textwidth]{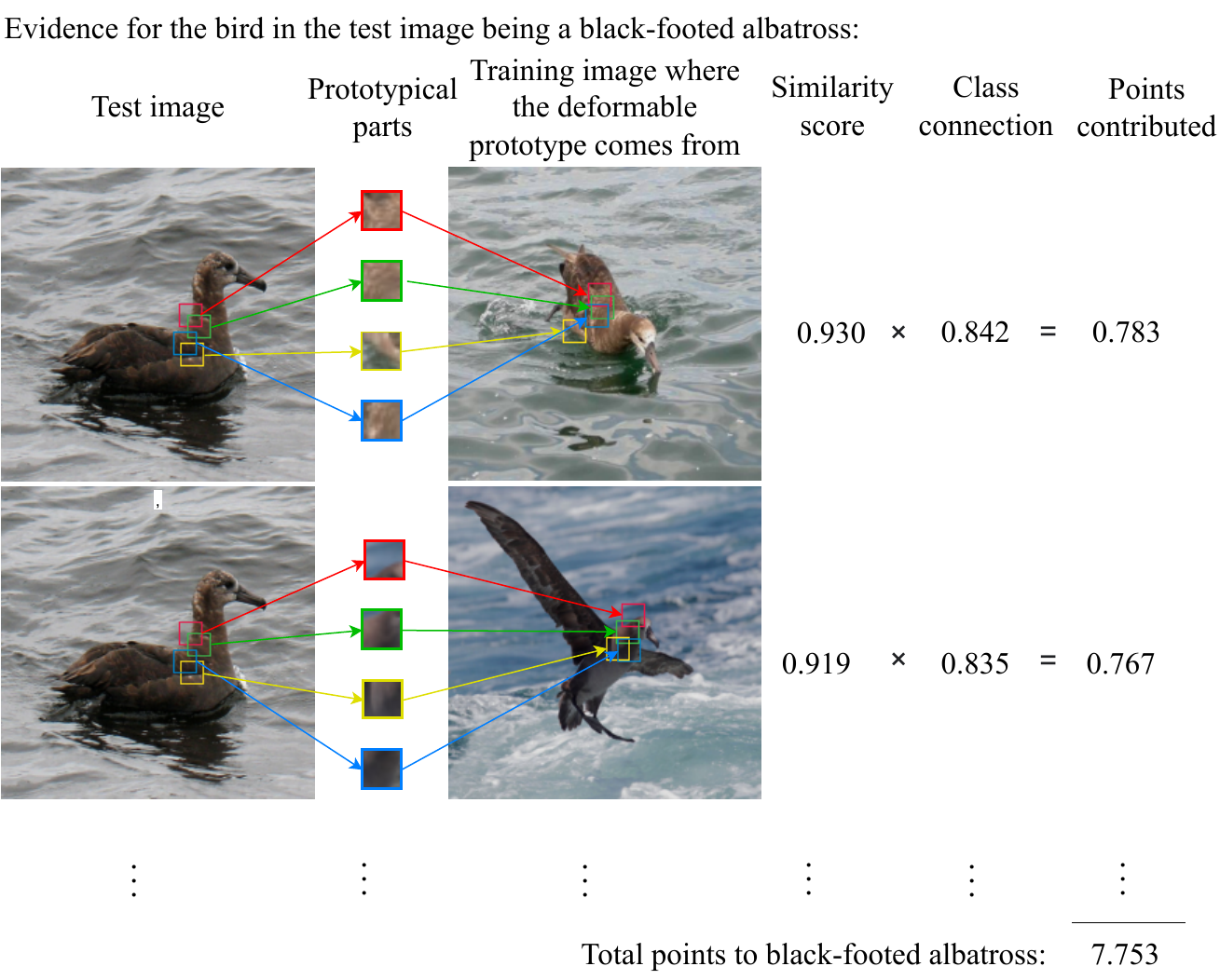}
    \includegraphics[width=0.47\textwidth]{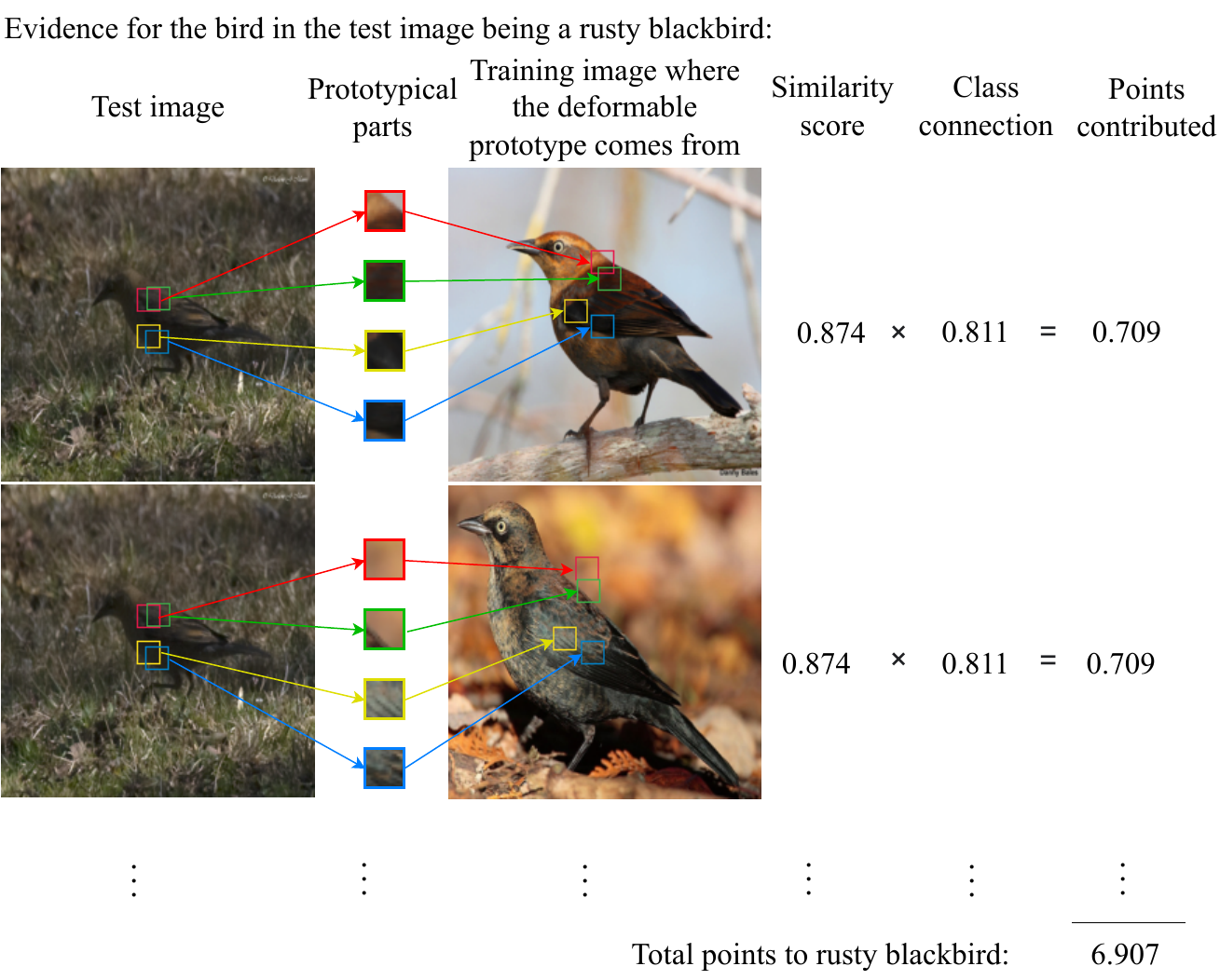}
    \includegraphics[width=0.47\textwidth]{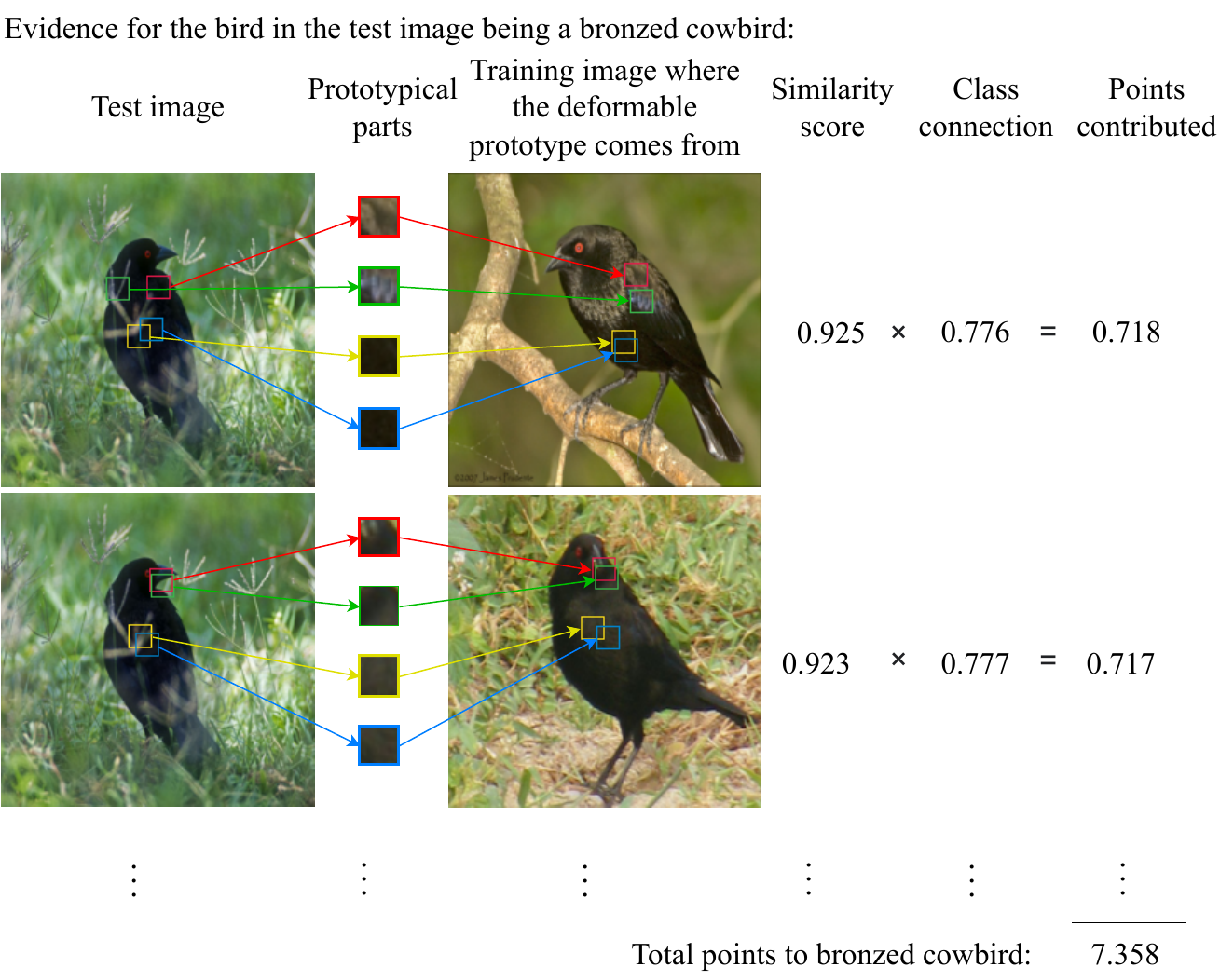}
  \caption{Example reasoning processes of a Deformable ProtoPNet with $2\times2$ prototypes when classifying a test image of a black-footed albatross (top), a rusty blackbird (middle), and a bronzed cowbird (bottom). In each case, we show the two deformable prototypes of the predicted class that produced the highest similarity scores.}
  \label{fig:reasoning_process_birds_2x2}
\end{figure}

\begin{figure}[t]
  \centering
    \includegraphics[width=0.47\textwidth]{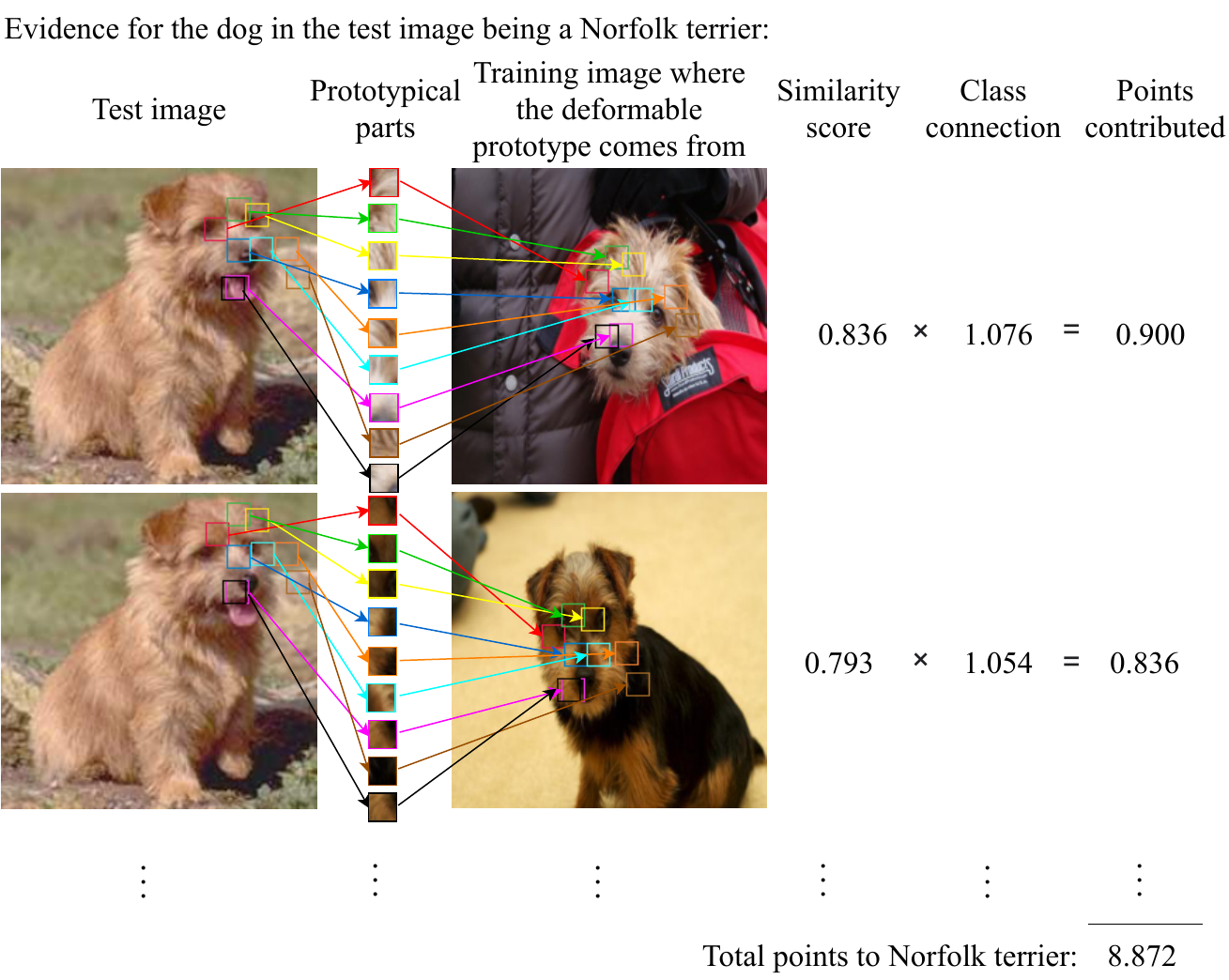}
    \includegraphics[width=0.47\textwidth]{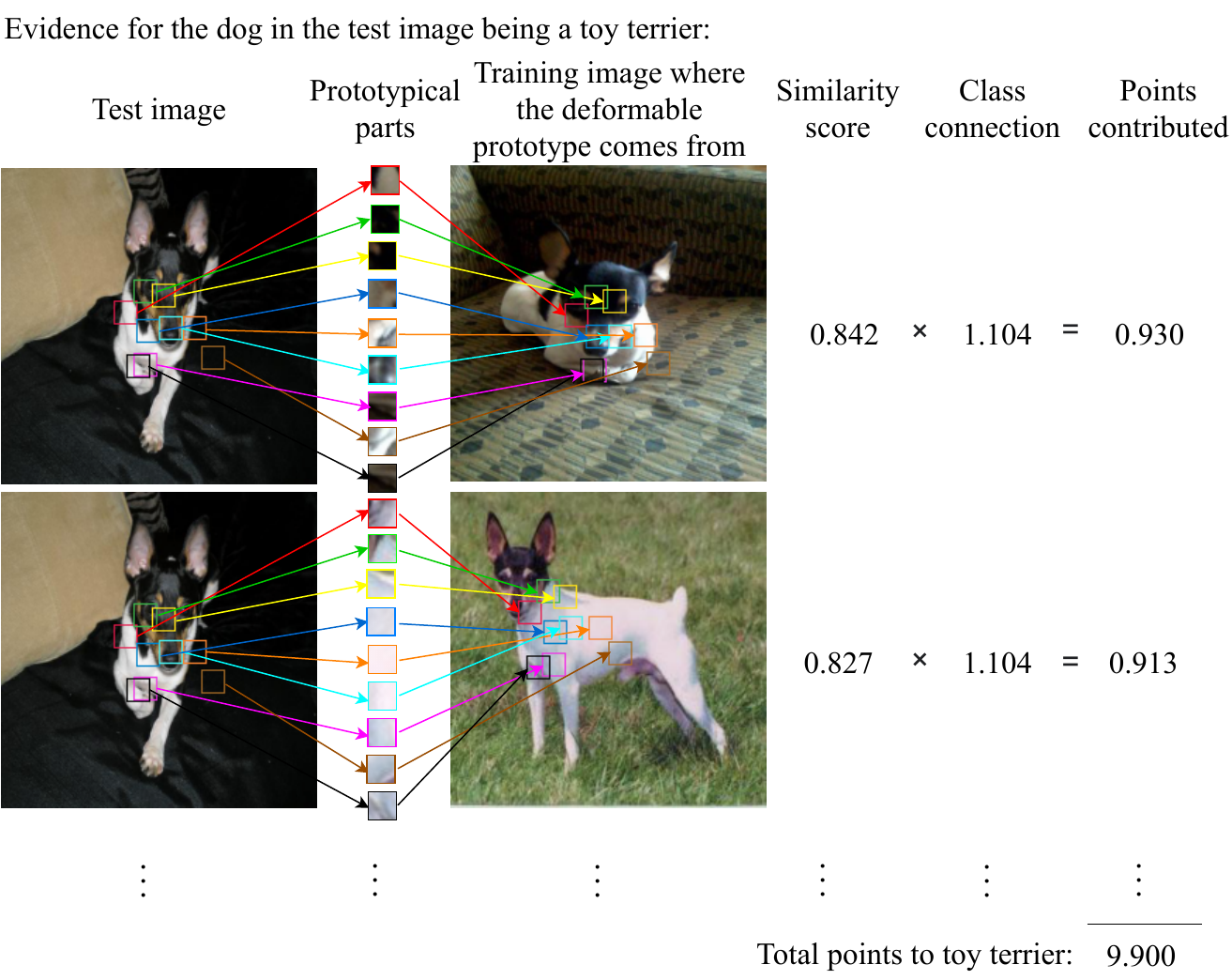}
    \includegraphics[width=0.47\textwidth]{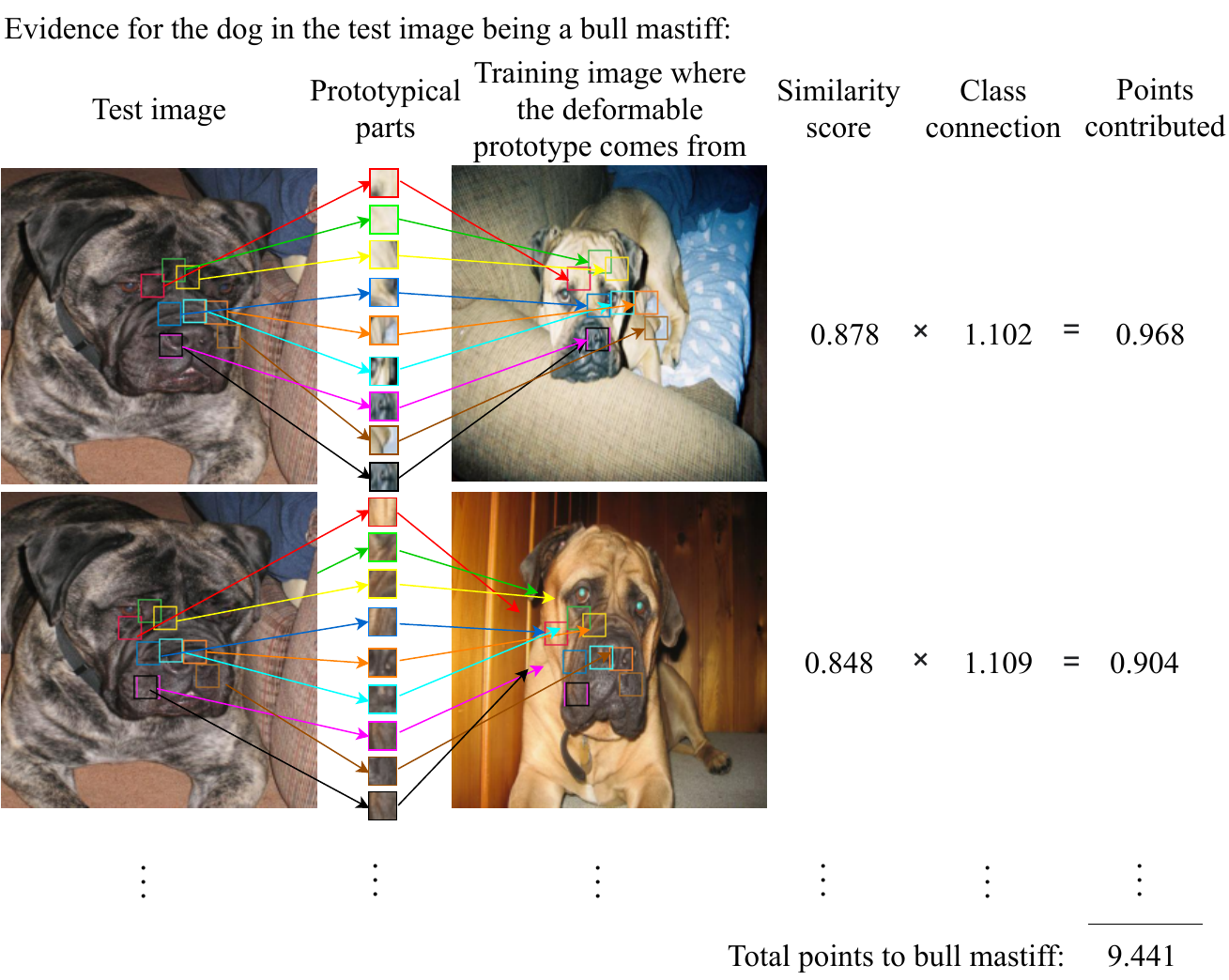}
  \caption{Example reasoning process of a Deformable ProtoPNet with $3\times3$ prototypes on a test image of a Norfolk terrier (top), a toy terrier (middle), and a bull mastiff (bottom). In each case, we show the two deformable prototypes of the predicted class that produced the highest similarity scores.}
  \label{fig:reasoning_process_dogs}
\end{figure}

We present more examples of reasoning processes produced by the top performing models on each dataset (for CUB-200-2011 \cite{cub_200}, we show ResNet50-based Deformable ProtoPNet \cite{resnet} with $3 \times 3$ prototypes and with $2 \times 2$ prototypes; for Stanford Dogs \cite{dogs}, we show Resnet152-based Deformable ProtoPNet with $3 \times 3$ prototypes) in Figure \ref{fig:reasoning_process_birds}, Figure \ref{fig:reasoning_process_birds_2x2} and Figure \ref{fig:reasoning_process_dogs}. In each case, we show the two deformable prototypes of the predicted class that produced the highest similarity scores. The similarity with these prototypes provide evidence for the test image belonging to the predicted class. For simplicity, we show only the evidence contributing to the predicted class for each test image.

Figure \ref{fig:reasoning_process_birds} shows the reasoning process of the best performing Deformable ProtoPNet on CUB-200-2011 \cite{cub_200} for a test image of an eastern towhee (top), a test image of an Acadian flycatcher (middle), and a test image of a pied billed grebe (bottom). To find evidence for the bird in Figure \ref{fig:reasoning_process_birds} (top) being an eastern towhee, our Deformable ProtoPNet compares each deformable prototype of the eastern towhee class to the test image by scanning the prototype across the test image (in the latent space of image features) -- in particular, the prototypical parts within a deformable prototype can adaptively change their relative spatial positions, as the deformable prototype moves across the test image (in the latent space), looking for image parts that are semantically similar to its prototypical parts. In the end, the Deformable ProtoPNet will take the highest similarity across the image for each deformable prototype as the similarity score between that prototype and the image -- for the test image of an eastern towhee in Figure \ref{fig:reasoning_process_birds} (top), the similarity score between the top prototype of the eastern towhee class and the test image is $0.960$. Since cosine similarity is used by a Deformable ProtoPNet, all similarity scores fall between $-1$ and $1$, so a similarity score of $0.960$ is very high. For each comparison with a deformable prototype, colored boxes on the test image in the figure show the spatial arrangement of the prototypical parts that is used to produce the similarity score. The similarity score produced by each deformable prototype is then multiplied by a class connection value to produce the points contributed by that prototype, and the points contributed by all prototypes within a class are added to produce a class score. The class with the highest class score is the predicted class. For the test image of an eastern towhee in Figure \ref{fig:reasoning_process_birds} (top), the class score of the eastern towhee class is $6.945$, which is the highest among all classes, making eastern towhee the predicted class.

Similarly, Figure \ref{fig:reasoning_process_birds} (middle) shows the reasoning process of the best performing Deformable ProtoPNet with $3 \times 3$ prototypes on CUB-200-2011 \cite{cub_200} for a test image of an Acadian flycatcher, and Figure \ref{fig:reasoning_process_birds} (bottom) shows the reasoning process for a test image of a pied billed grebe.

Figure \ref{fig:reasoning_process_birds_2x2} shows the reasoning process of the best performing Deformable ProtoPNet with $2 \times 2$ prototypes on CUB-200-2011 \cite{cub_200} for a test image of a black-footed albatross (top), a test image of a rusty blackbird (middle) and for a test image of a bronzed cowbird (bottom).

Figure \ref{fig:reasoning_process_dogs} shows the reasoning process of the best performing Deformable ProtoPNet on Stanford Dogs \cite{dogs} for a test image of a Norfolk terrier (top), a test image of a toy terrier (middle) and for a test image of a bull mastiff (bottom).

\section{Local Analysis: Visualizations of Most Similar Prototypes to Given Images}


\begin{figure}[t]
  \centering
    \includegraphics[width=0.47\textwidth]{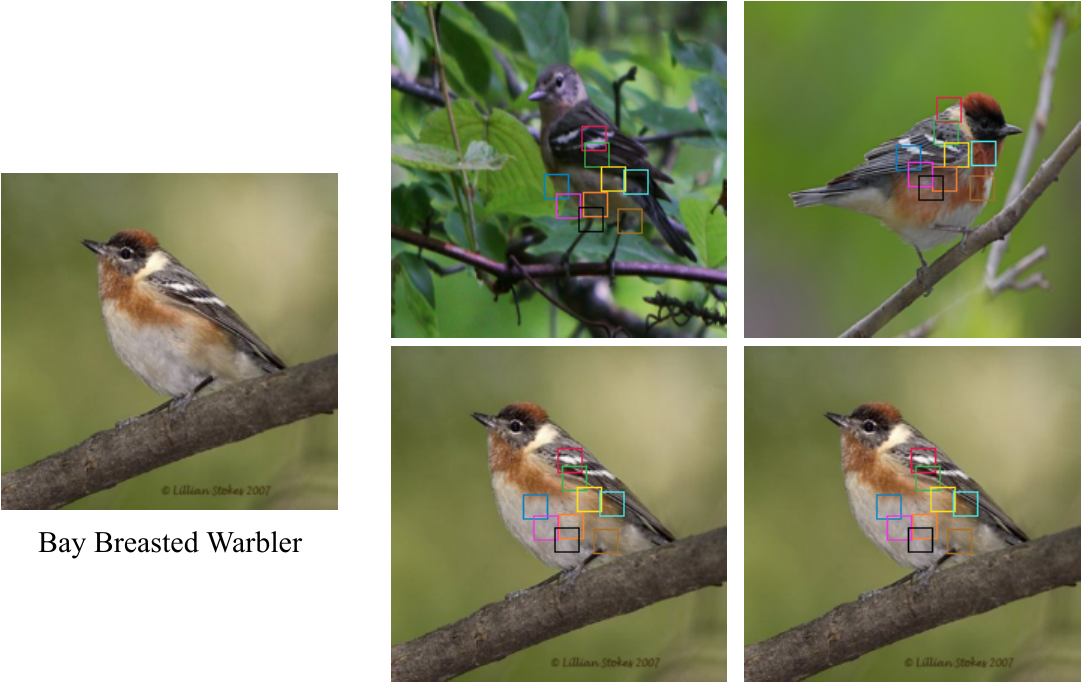}
    \includegraphics[width=0.47\textwidth]{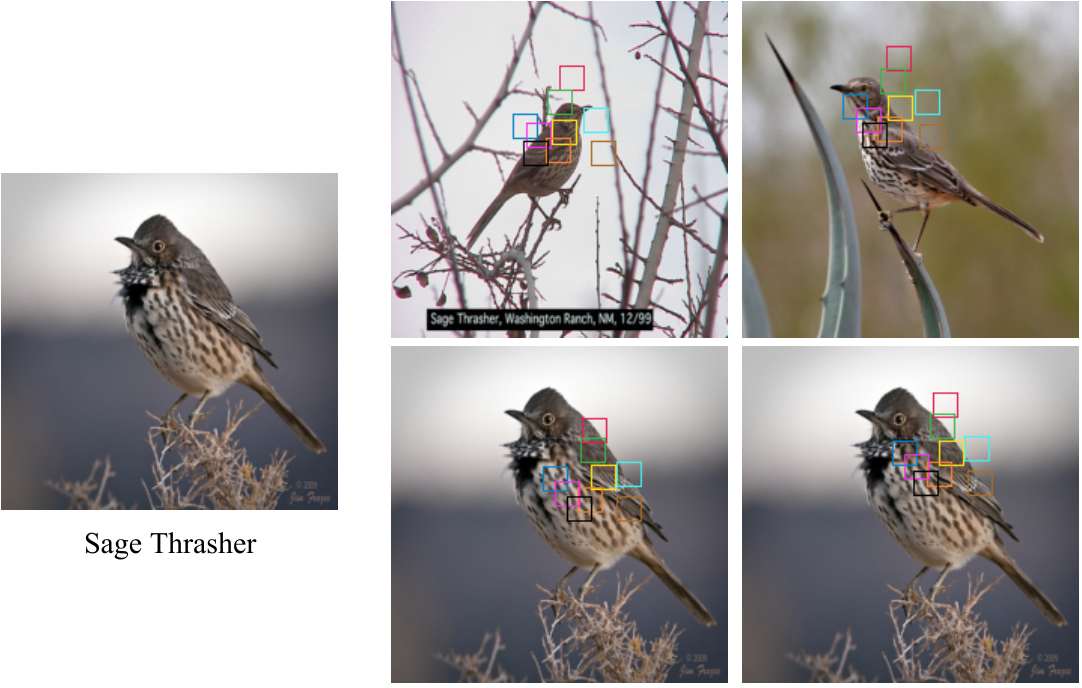}
    \includegraphics[width=0.47\textwidth]{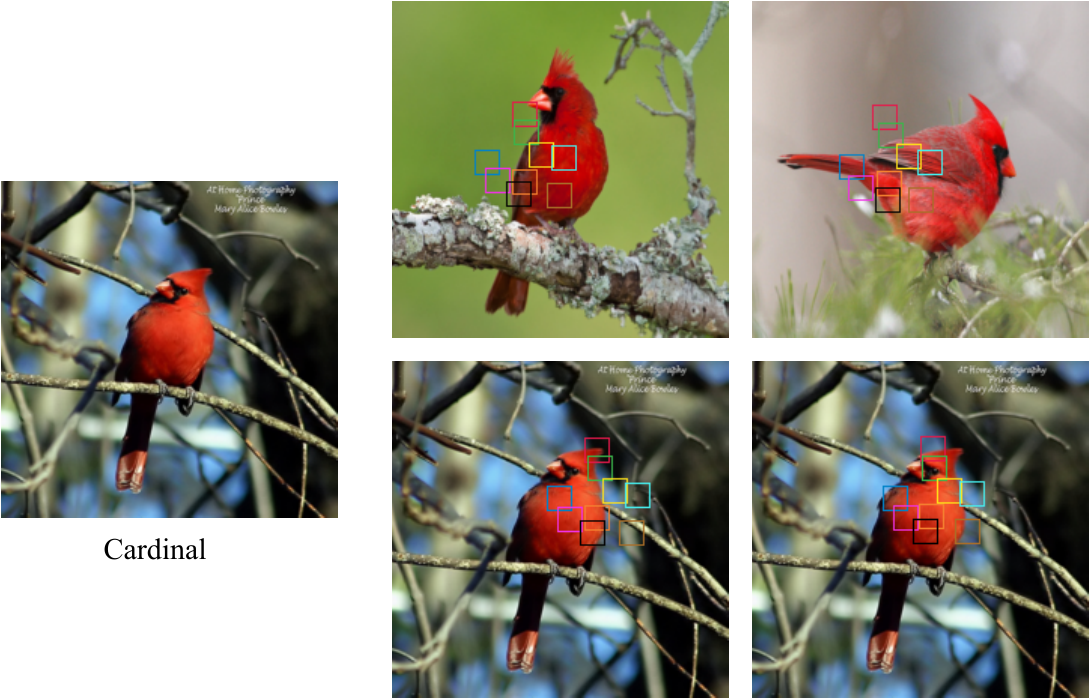}
  \caption{Local analyses of three test images from CUB-200-2011 \cite{cub_200} for a Deformable ProtoPNet with $3\times 3$ prototypes. For each test image on the left, the top row shows the two most similar deformable prototypes, and the bottom row shows the spatial arrangement of the prototypical parts on the test image that produced the similarity score for the corresponding prototype.}
  \label{fig:local_analysis_birds}
\end{figure}

\begin{figure}[t]
  \centering
    \includegraphics[width=0.47\textwidth]{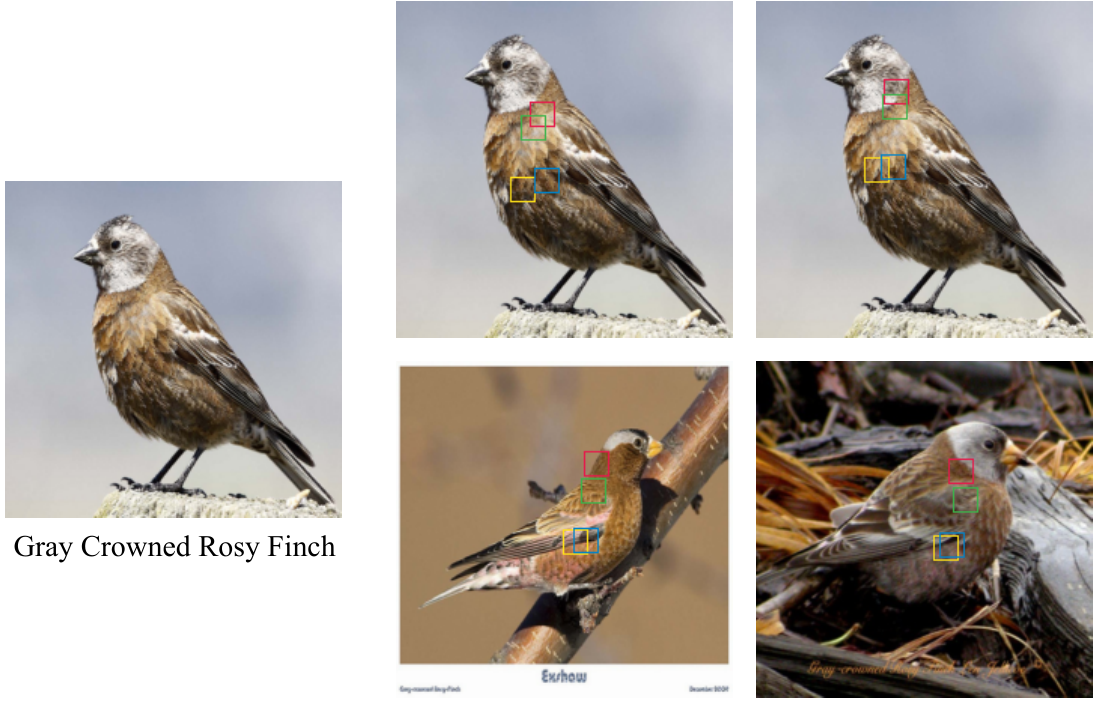}
    \includegraphics[width=0.47\textwidth]{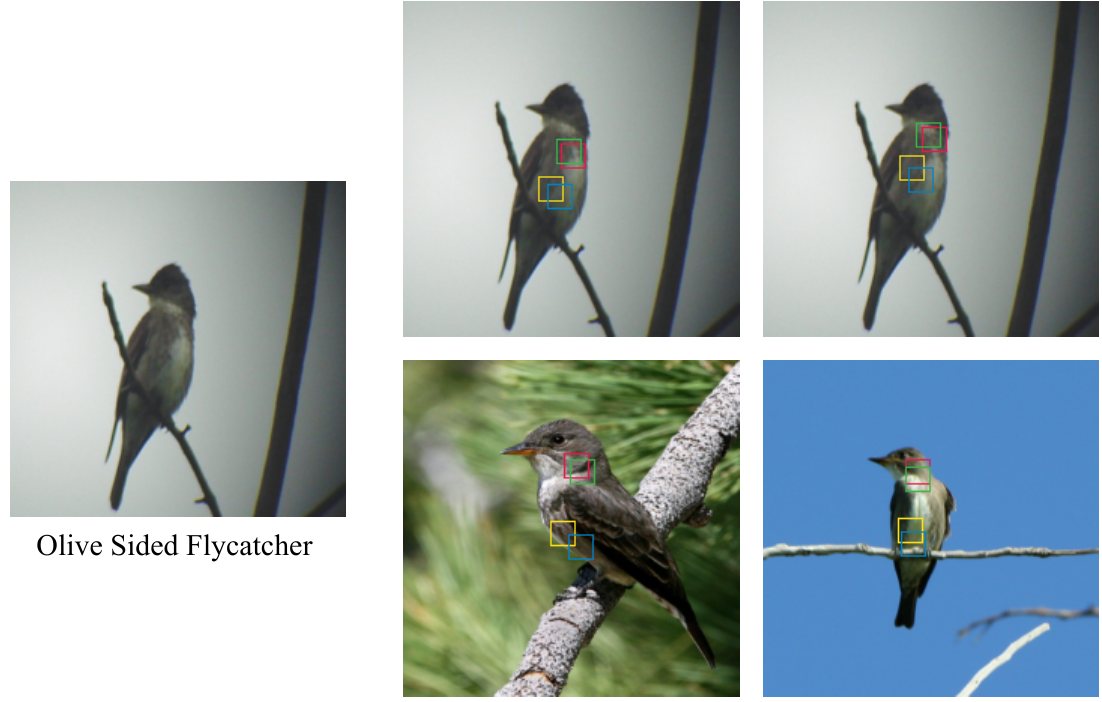}
    \includegraphics[width=0.47\textwidth]{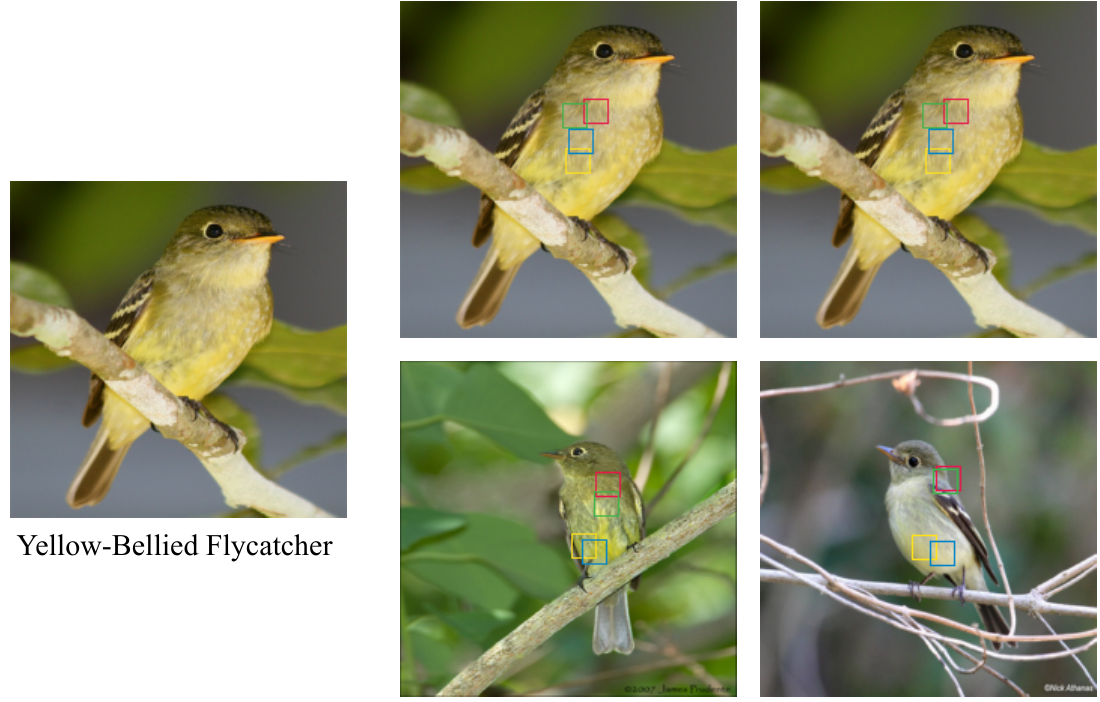}
  \caption{Local analyses of three test images from CUB-200-2011 \cite{cub_200} for a Deformable ProtoPNet with $2\times 2$ prototypes. For each test image on the left, the top row shows the two most similar deformable prototypes, and the bottom row shows the spatial arrangement of the prototypical parts on the test image that produced the similarity score for the corresponding prototype.}
  \label{fig:local_analysis_birds_2x2}
\end{figure}

\begin{figure}[t]
    \centering
    \includegraphics[width=0.47\textwidth]{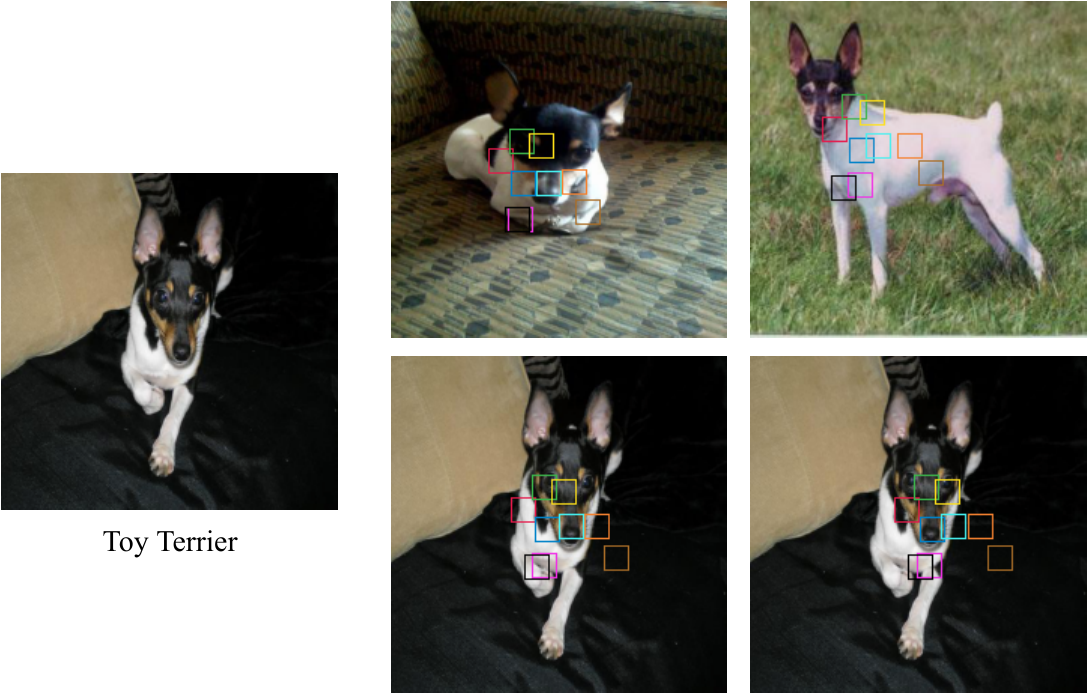}
    \includegraphics[width=0.47\textwidth]{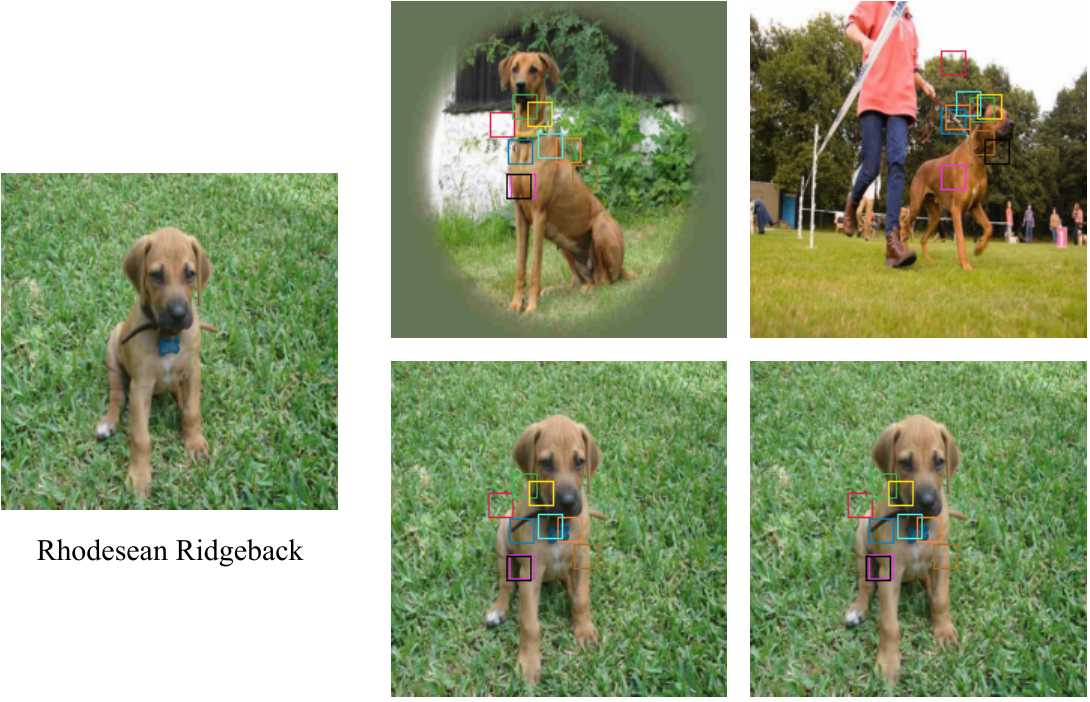}
    \includegraphics[width=0.47\textwidth]{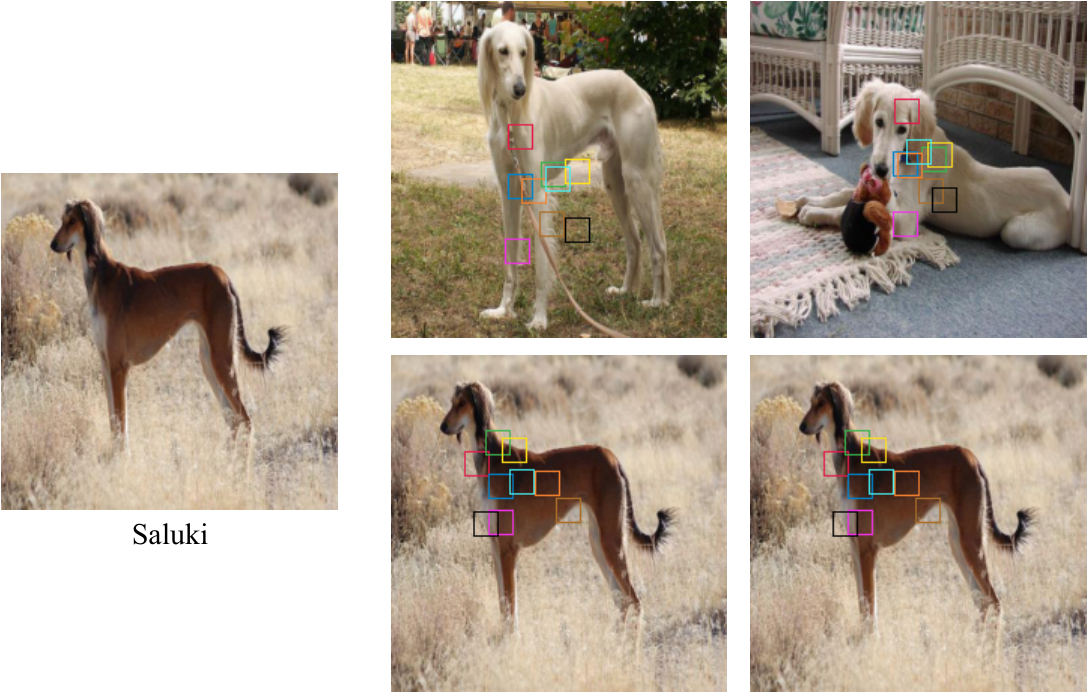}
    \caption{Local analyses of three test images from Stanford Dogs \cite{dogs} for a Deformable ProtoPNet with $3\times 3$ prototypes. For each test image on the left, the top row shows the two most similar deformable prototypes, and the bottom row shows the spatial arrangement of the prototypical parts on the test image that produced the similarity score for the corresponding prototype.}
    \label{fig:local_analysis_dogs}
\end{figure}

In this section, we visualize the most similar prototypes to a given test image (we call this a \textit{local analysis} of the test image), for a number of test images. Figure \ref{fig:local_analysis_birds} shows the two most similar deformable prototypes (learned by the best performing Deformable ProtoPNet using $3\times3$ prototypes) for each of three test images from CUB-200-2011 \cite{cub_200}. Figure \ref{fig:local_analysis_birds_2x2} shows the two most similar deformable prototypes (learned by the best performing Deformable ProtoPNet using $2\times2$ prototypes) for each of three test images from CUB-200-2011 \cite{cub_200}. Figure \ref{fig:local_analysis_dogs} shows the two most similar deformable prototypes (learned by the best performing Deformable ProtoPNet) for each of three test images from Stanford Dogs \cite{dogs}. For each test image on the left, the top row shows the two most similar deformable prototypes, and the bottom row shows the spatial arrangement of the prototypical parts on the test image that produced the similarity score for the corresponding prototype. In general, the most similar prototypes for a given image come from the same class as that of the image, and there is some semantic correspondence between a prototypical part and the image patch it is compared to under the spatial arrangement of the prototypical parts where the deformable prototype achieves the highest similarity across the image.

\section{Global Analysis: Visualizations of Most Similar Images to Given Prototypes}


\begin{figure*}[t]
  \centering
    \includegraphics[height=0.14\textheight]{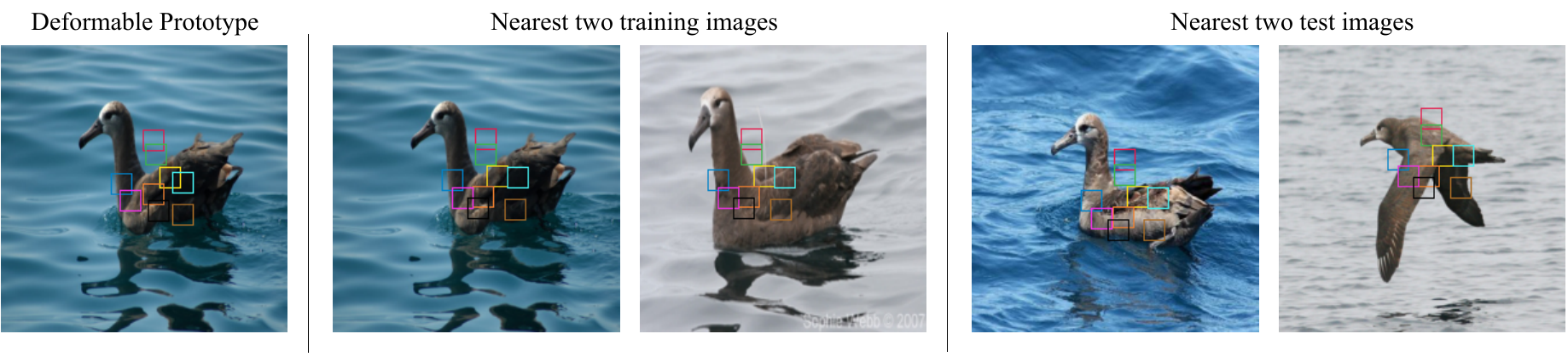}
    \includegraphics[height=0.14\textheight]{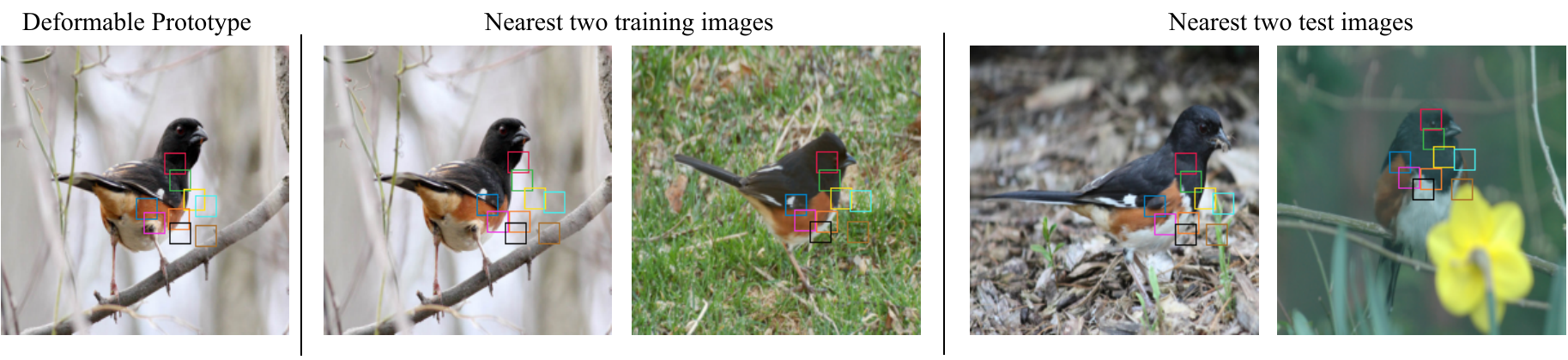}
    \includegraphics[height=0.14\textheight]{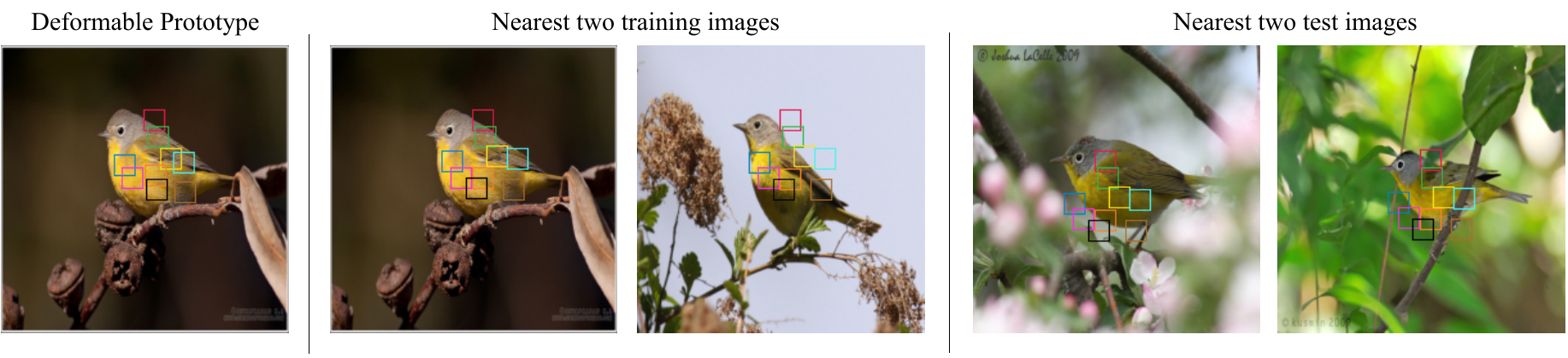}
  \caption{Global analyses of three deformable prototypes from CUB-200-2011 \cite{cub_200} for a Deformable ProtoPNet with  $3 \times 3$ prototypes. For each deformable prototype on the left, we show two most similar images from the training set (middle) and two most similar images from the test set (right).}
  \label{fig:global_analysis_birds}
\end{figure*}

\begin{figure*}[t]
  \centering
    \includegraphics[height=0.14\textheight]{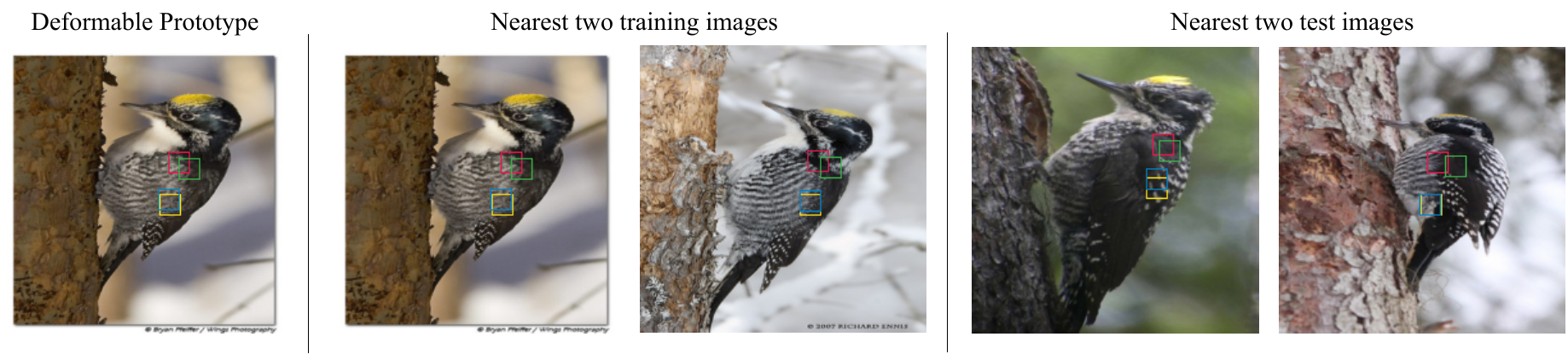}
    \includegraphics[height=0.14\textheight]{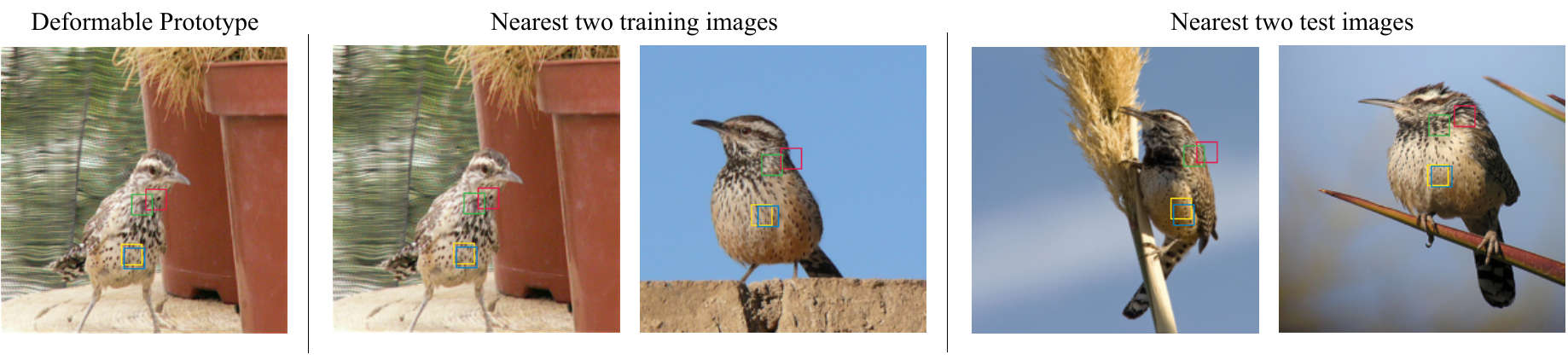}
    \includegraphics[height=0.14\textheight]{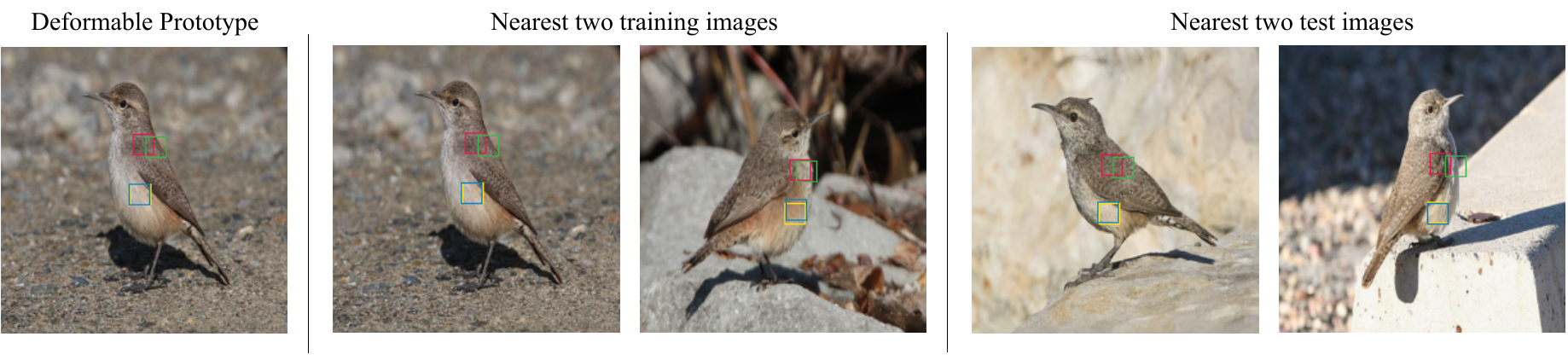}
  \caption{Global analyses of three deformable prototypes from CUB-200-2011 \cite{cub_200} for a Deformable ProtoPNet with $2 \times 2$ prototypes. For each deformable prototype on the left, we show two most similar images from the training set (middle) and two most similar images from the test set (right).}
  \label{fig:global_analysis_birds_2x2}
\end{figure*}

\begin{figure*}[t]
  \centering
    \includegraphics[height=0.14\textheight]{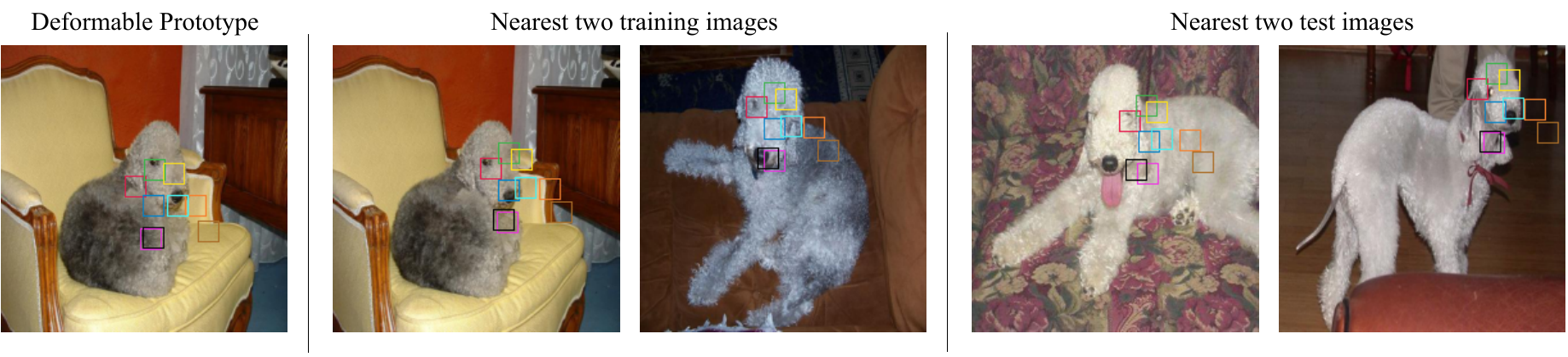}
    \includegraphics[height=0.14\textheight]{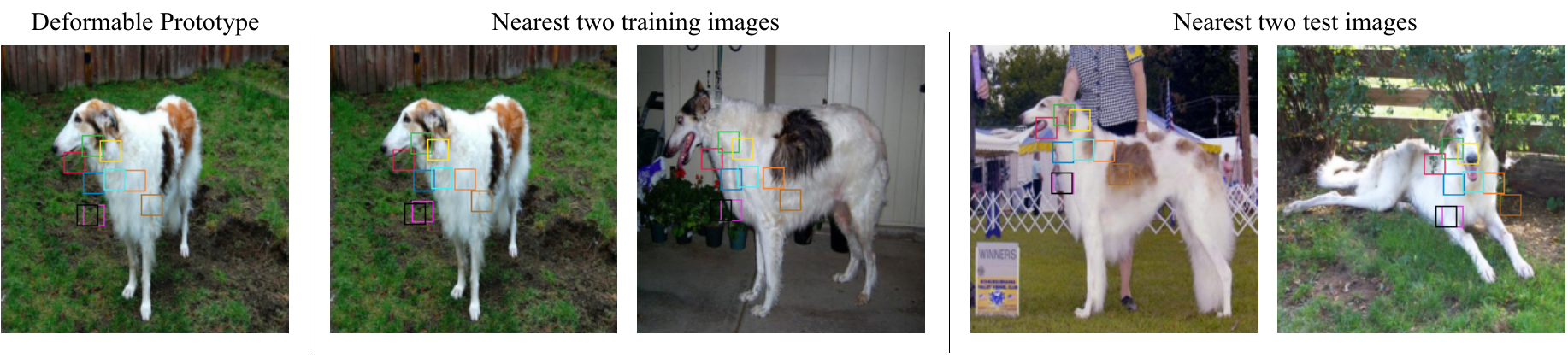}
    \includegraphics[height=0.14\textheight]{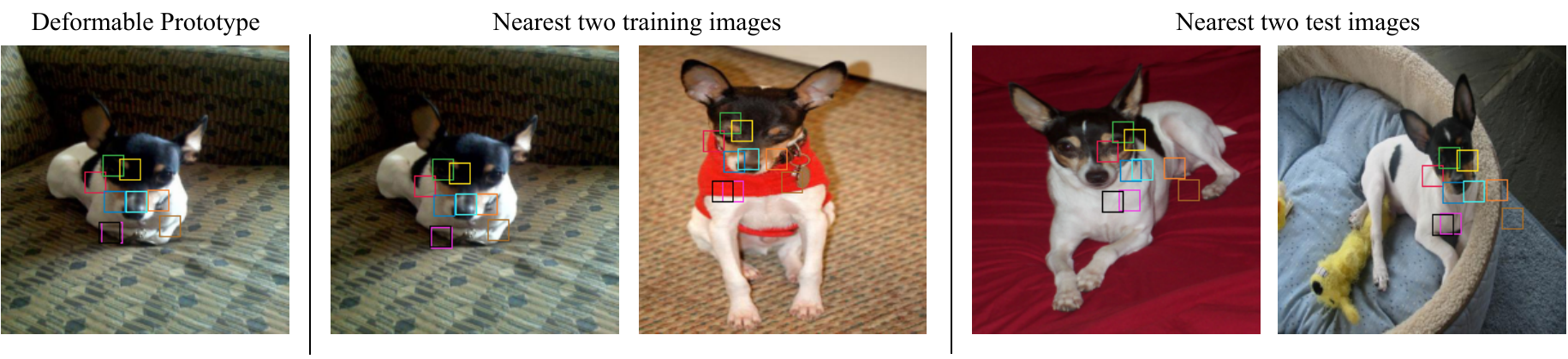}
  \caption{Global analysis for three deformable prototypes from Stanford Dogs \cite{dogs} for a Deformable ProtoPNet with  $3 \times 3$ prototypes. For each deformable prototype on the left, we show two most similar images from the training set (middle) and two most similar images from the test set (right).}
  \label{fig:global_analysis_dogs}
\end{figure*}

In this section, we visualize the most similar training and test images to a given deformable prototype (we call this a \textit{global analysis} of the prototype), for a number of deformable prototypes. Figure \ref{fig:global_analysis_birds} shows the two most similar training images and the two most similar test images for each of three deformable prototypes learned by the best performing Deformable ProtoPNet with $3\times3$ prototypes on CUB-200-2011 \cite{cub_200}. Figure \ref{fig:global_analysis_birds_2x2} shows the two most similar training images and the two most similar test images for each of three deformable prototypes learned by the best performing Deformable ProtoPNet with $3\times3$ prototypes on CUB-200-2011 \cite{cub_200}. Figure \ref{fig:global_analysis_dogs} shows the two most similar training images and the two most similar test images for each of three deformable prototypes learned by the best performing Deformable ProtoPNet on Stanford Dogs \cite{dogs}. For each deformable prototype on the left, we show two most similar images from the training set (middle) and two most similar images from the test set (right). In general, the most similar training and test images for a given deformable prototype come from the same class as that of the prototype, and for each of the most similar images, there is some semantic correspondence between most prototypical parts and the image patches they are compared to under the spatial arrangement of the prototypical parts where the deformable prototype achieves the highest similarity across the image.

\section{Numerical Results on Stanford Dogs}


\begin{table*}[h]
    \centering
    \begin{tabular}{|p{0.19\textwidth}|c|c|c|}
    \hline
    Method & VGG-19 & ResNet-152 & Densenet161 \\ 
    \hline
    Baseline & 77.3 & 85.2 & \textbf{84.1} \\ \hline 
    ProtoPNet \cite{ProtoPNet} & 73.6 & 76.2 & 77.3 \\ \hline 
    Def. ProtoPNet (nd) & 74.8 & \textbf{86.5} & 83.7  \\ \hline
    Def. ProtoPNet & \textbf{77.9} & \textbf{86.5} & 83.7 \\ \hline
    \end{tabular}
    \caption{Accuracy of Deformable ProtoPNet compared to the baseline model, ProtoPNet \cite{ProtoPNet}, and Deformable ProtoPNet without deformations (denoted (nd)) across different base architectures on the Stanford Dogs dataset \cite{dogs}.}
    \label{tab:dogs_numerical}
\end{table*}

\begin{table}[h]
    \centering
    \begin{tabular}{|l|l|}
    \hline
    Interpretability & Model: accuracy              \\ \hline
    
    \begin{tabular}[c]{@{}l@{}}Part-level\\ attention \end{tabular}                                                                            & \begin{tabular}[c]{@{}l@{}} FCAN\cite{FCAN}: 84.2\\ 
    \textbf{RA-CNN}\cite{fu2017RACNN}: \textbf{87.3}\end{tabular} \\ \hline
    
    \begin{tabular}[c]{@{}l@{}}Part-level attn. +\\ learned prototypes\end{tabular} &
    
    \begin{tabular}[c]{@{}l@{}} 
        ProtoPNet\cite{ProtoPNet}: 77.3\\
        \textbf{Def. ProtoPNet(nd)}: \textbf{86.5}
    \end{tabular} \\
    \hline
    
    \begin{tabular}[c]{@{}l@{}}Part-level attn. +\\ learned prototypes +\\ deformations\end{tabular} &
    
    \begin{tabular}[c]{@{}l@{}} 
        \textbf{Def. ProtoPNet}: \textbf{86.5}
    \end{tabular}
    
    
    
    \\ \hline
    \end{tabular}
    \caption{Accuracy and interpretability of Deformable ProtoPNet compared to other interpretable models on Stanford Dogs \cite{dogs}. We use (nd) to indicate a Deformable ProtoPNet without using deformations.}
    \label{tab:supp:state-of-the-art}
\end{table}

We conducted another case study of our Deformable ProtoPNet on Stanford Dogs \cite{dogs}. We trained each Deformable ProtoPNet with 10 $3 \times 3$ deformable prototypes per class, where each prototype was composed of 9 prototypical parts. We ran experiments using VGG-19 \cite{vgg}, ResNet-152 \cite{resnet}, and DenseNet-161 \cite{densenet} as CNN backbones. All backbones were pretrained using ImageNet \cite{deng2009imagenet}.

\textbf{We find that Deformable ProtoPNet can achieve competitive accuracy across multiple backbone architectures}. As shown in Table \ref{tab:dogs_numerical}, our Deformable ProtoPNet achieves higher accuracy than the baseline uninterpretable architecture in two out of three cases, including the highest performing model based on ResNet-152 \cite{resnet}. In all cases, we achieve substantially higher accuracy than ProtoPNet \cite{ProtoPNet}.

\textbf{We find that using deformations improves or maintains accuracy}. As shown in Table \ref{tab:dogs_numerical}, a Deformable ProtoPNet with deformations achieves a level of accuracy higher than or equal to the corresponding model without deformations across all three CNN backbones -- in particular, a Deformable ProtoPNet with deformations achieves a test accuracy more than $3\%$ higher than the one without deformations when VGG-19 is used as a backbone.

\textbf{We find that a single Deformable ProtoPNet achieves accuracy on par with the state-of-the-art.} As Table \ref{tab:supp:state-of-the-art} shows, a Deformable ProtoPNet can achieve accuracy ($86.5\%$) competitive with the state-of-the-art.

\section{Experimental Setup}

\subsection{Hyperparameters}

We ran experiments using VGG \cite{vgg}, ResNet \cite{resnet}, and DenseNet \cite{densenet} as CNN backbones $f$. The ResNet-50 backbone was pretrained on iNaturalist \cite{inaturalist_2018}, and all other backbones were pretrained using ImageNet \cite{deng2009imagenet}. We used $14 \times 14$ as the spatial dimension (height and width) of the latent image-feature tensor. In particular, since the CNN backbones produce feature maps of spatial dimension $7 \times 7$, we obtain $14 \times 14$ feature maps by removing the final max pooling from the backbone architecture, or by upsampling from $7 \times 7$ feature maps via bilinear interpolation. The convolutional feature maps are augmented with a uniform channel of value $\epsilon = 10^{-5}$, and then normalized and rescaled at each spatial position as described in the main paper.

When prototypes were allowed to deform, we used two convolutional layers to predict offsets for prototype deformations -- the first convolutional layer has $128$ output channels, and the second convolutional layer has either $18$ output channels for $3 \times 3$ prototypes (to produce $2$ offsets for each of the $9$ prototypical parts at each spatial position) or $8$ output channels for $2 \times 2$ prototypes (to produce $2$ offsets for each of the $4$ prototypical parts at each spatial position).

For our experiments on CUB-200-2011 \cite{cub_200}, we trained each Deformable ProtoPNet with 6 deformable prototypes per class when each prototype was composed of 9 prototypical parts, and 10 deformable prototypes per class when each prototype was composed of 4 prototypical parts (except where otherwise specified). For our experiments on Stanford Dogs \cite{dogs}, we trained each Deformable ProtoPNet with 10 deformable prototypes per class where each prototype was composed of 9 prototypical parts.

We trained each Deformable ProtoPNet for 30 epochs. In particular, we started our training with a ``warm-up'' stage, in which we loaded and froze the pre-trained weights and biases and we froze the offset prediction branch, and focused on training the deformable prototype layer for 5 epochs (7 epochs for VGG- and DenseNet-based Deformable ProtoPNets on CUB-200-2011 \cite{cub_200}), at a learning rate of $3 \times 10^{-3}$. We then performed a second ``warm-up'' stage, in which we froze the offset prediction branch and focused on training the deformable prototype layer as well as the weights and biases of the CNN backbone for 5 more epochs, at a learning rate of $1 \times 10^{-4}$ for the CNN backbone parameters and $3 \times 10^{-3}$ for the deformable prototypes. Finally, we jointly trained all model parameters for the remaining training epochs, at a starting learning rate of $1 \times 10^{-4}$ for the CNN backbone parameters, $3 \times 10^{-3}$ for the deformable prototypes, and $5 \times 10^{-4}$ for the offset prediction branch. We reduced the learning rate by a factor of 0.1 every 5 epochs. We performed prototype projection and last layer optimization at epoch 20 and epoch 30.

\subsection{Hardware and Software}

Experiments were conducted on two types of servers. The first type of servers comes with Intel Xeon E5-2620 v3 (2.4GHz) CPUs and two to four NVIDIA A100 SXM4 40GB GPUs, and the experiments were run on the first type of servers with PyTorch version 1.8.1 and CUDA version 11.1. The second type of servers comes with TensorEX TS2-673917-DPN Intel Xeon Gold 6226 Processor (2.7Ghz) CPUs with two NVIDIA Tesla 2080 RTX Ti GPUs, and the experiments were run on the second type of servers with PyTorch version 1.10.0 and CUDA version 10.2.

The deformable prototypes were implemented in CUDA C++, while other components of Deformable ProtoPNet were implemented in Python 3 using PyTorch. Code is available at \href{https://github.com/jdonnelly36/Deformable-ProtoPNet}{https://github.com/jdonnelly36/Deformable-ProtoPNet}.

\FloatBarrier

\end{document}